\documentclass[10pt,letterpaper]{article}

\usepackage[letterpaper,left=1.06in,right=1.06in,top=.83in,bottom=.89in,footskip=30pt]{geometry}
\usepackage{times,amsmath,amssymb,graphicx,booktabs,tabularx,xcolor}
\usepackage[format=plain,labelformat=simple,labelsep=period,font=small,compatibility=false]{caption}
\usepackage{authblk}
\usepackage[numbers,sort&compress]{natbib}
\usepackage[hyphens]{url}
\definecolor{linkblue}{rgb}{.13,.28,.42}
\usepackage[breaklinks,colorlinks,allcolors=linkblue,pdfencoding=auto]{hyperref}
\graphicspath{{fig/}}
\let\AppendixSection\section
\let\AppendixSubsection\subsection
\let\AppendixSubsubsection\subsubsection
\hypersetup{pdftitle={Semantic State Reading for Joint Image-Semantic Diffusion Transformers},
  pdfauthor={Guanxi Lai, Zixuan Li, Zhiwei Li, Qi Li}}
\title{Semantic State Reading for Joint\\Image--Semantic Diffusion Transformers}

\renewcommand{\Affilfont}{\small}
\author[1]{Guanxi Lai}
\author[2]{Zixuan Li}
\author[3,4]{Zhiwei Li}
\author[4]{Qi Li}
\makeatletter
\affil[1]{Tsinghua University}
\def\AB@affilsep{\protect\space\protect\quad\protect\Affilfont}
\affil[2]{University of International Business and Economics}
\affil[3]{University of the Chinese Academy of Sciences}
\def\AB@affilsep{\protect\space\protect\quad\protect\Affilfont}
\affil[4]{Institute of Automation, Chinese Academy of Sciences}
\makeatother
\date{}
\makeatletter
\renewcommand{\maketitle}{
  \begin{center}
    {\fontsize{21}{24}\selectfont\bfseries\@title\par}
    \vspace{5pt}
    {\large\begin{tabular}{c}\@author\end{tabular}\par}
  \end{center}}
\makeatother
\renewenvironment{abstract}
  {\begin{list}{}{\setlength\leftmargin{.22in}\setlength\rightmargin{.22in}
     \setlength\topsep{4pt}\setlength\parsep{0pt}}\item[]
   {\centering\normalsize\bfseries\abstractname\par}\vspace{5pt}
   \fontsize{10}{12.6}\selectfont\noindent\ignorespaces}
  {\end{list}}
\begin{document}
\begingroup
\fontsize{10.95}{14}\selectfont
\raggedbottom
\setlength{\emergencystretch}{1.9em}
\setlength{\parindent}{1.15em}
\setlength{\parskip}{0pt plus .25pt}
\setlength{\bibsep}{1.5pt plus .4pt}
\setlength{\textfloatsep}{13pt plus 2pt minus 2pt}
\setlength{\floatsep}{12pt plus 2pt minus 2pt}
\setlength{\intextsep}{11pt plus 2pt minus 2pt}
\renewcommand{\topfraction}{.88}
\renewcommand{\bottomfraction}{.76}
\renewcommand{\textfraction}{.10}
\renewcommand{\floatpagefraction}{.78}
\setcounter{topnumber}{3}
\setcounter{bottomnumber}{2}
\setcounter{totalnumber}{4}
\captionsetup{labelfont=bf,justification=raggedright,singlelinecheck=false,skip=5pt}
\captionsetup[table]{position=above}
\maketitle
\begin{center}
\centering\includegraphics[width=\textwidth]{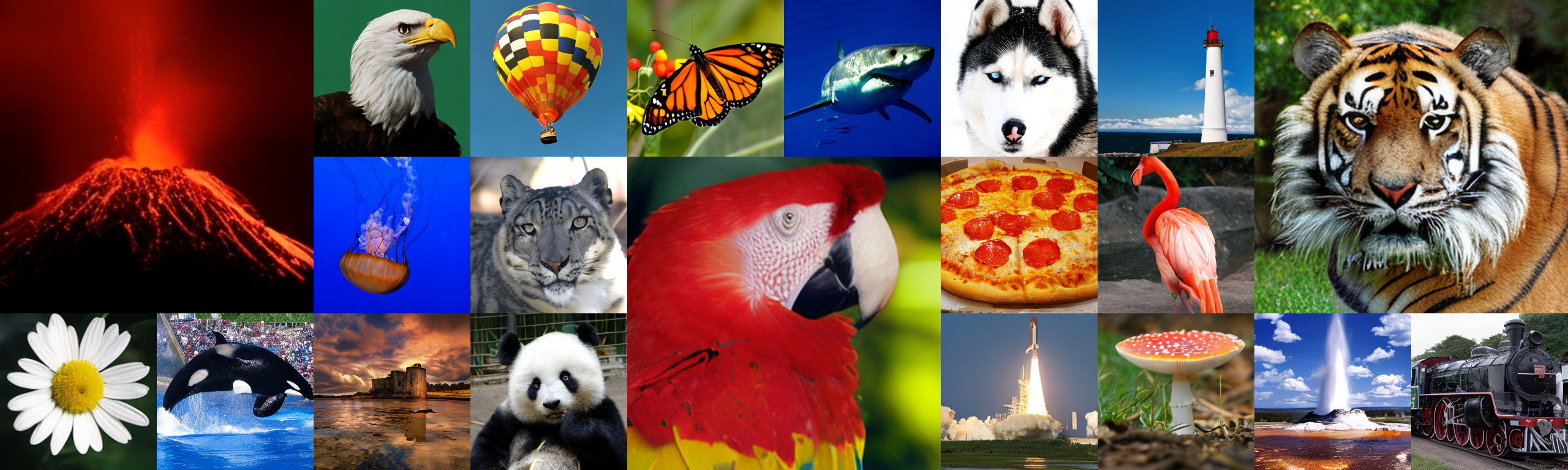}
\captionsetup{hypcap=false}
\captionof{figure}{\textbf{Class-conditional image samples generated with Semantic State Reading (SSR).}}
\label{fig:showcase}
\end{center}

\begin{abstract}
Pretrained visual representations improve diffusion-transformer training, yet exploiting both global context and local spatial structure requires deciding which semantic states to model and how to access them. We introduce Semantic State Reading (SSR), which jointly denoises image latents, a global CLS token encoding image-level context, and a compact patch state preserving spatial correspondence. SSR couples this global--local modeling with distinct reading paths: local states are fused with corresponding image tokens at the input, while independent projections inject the current global state into every transformer block. Both semantic scopes thus remain accessible throughout denoising. Ablations show complementary benefits from local-state modeling and repeated global reading, and semantic-trajectory interventions alter image layout under a fixed class label. On ImageNet $256\times256$, SiT-XL/2 with SSR reaches or surpasses the reference FIDs of SiT, REPA, and Representation Entanglement for Generation (REG) with 100$\times$, 40$\times$, and 2$\times$ fewer training iterations, respectively. SSR achieves unguided FID 1.78 at 2M iterations and FID 1.34 with classifier-free guidance alone. Combining classifier-free guidance with autoguidance yields FID 1.31 at 1M iterations, achieving the best FID in our main comparison table.
\end{abstract}

\section{Introduction}
\label{sec:intro}
Diffusion transformers have become an important architecture for high-quality image generation~\cite{dit,uvit,pixart,sd3}. Latent diffusion~\cite{ldm} reduces image-modeling costs through a pretrained autoencoder, DiT~\cite{dit} introduces a scalable transformer backbone, and SiT~\cite{sit} studies generative paths and sampling through stochastic interpolants. However, these models must still learn complex semantic structure through denoising~\cite{ddpm,iddpm,mdt}, and high-quality generation often requires millions of training updates. Pretrained visual representations provide semantic priors that can ease this learning burden and accelerate convergence~\cite{repa,maskalign,rae}.

Existing methods primarily use these representations through alignment or joint generation. REPA~\cite{repa} aligns intermediate generator features with clean teacher representations, while MaskAlign~\cite{maskalign} improves supervision through token mixing and subset alignment and incorporates global-token modeling. Joint-generation methods further introduce semantics into the diffusion state: Representation Entanglement for Generation (REG)~\cite{reg} jointly denoises image latents and a global CLS token, ReDi~\cite{redi} introduces PCA-compressed spatial patch features, and REGLUE~\cite{reglue} combines a global token with compressed multi-layer spatial features. These studies show that semantics can both supervise training and be generated alongside images.

However, incorporating semantics into generation does not by itself fully exploit their spatial structure or determine how they are accessed. A global CLS token provides image-level context without explicit patch-level spatial correspondence; local patch features describe position-specific content but do not directly replace an image-level summary~\cite{dino,dinov2,vit}. REG~\cite{reg} and ReDi~\cite{redi} emphasize one type of state, while REGLUE~\cite{reglue} demonstrates the value of joint modeling. Building on these findings, we ask two connected questions: how should global and local semantics be combined, and how should the network access them throughout denoising?

Representation alignment constrains what intermediate features learn, but does not specify how each layer reads the current semantic state. Similarly, after input-level semantic fusion, deeper blocks primarily access semantics through progressively transformed hidden features, a different computational path from directly reading the current state. Joint modeling therefore determines which semantic information is available, while the reading design determines how that information participates in deeper denoising computations.

We introduce \emph{Semantic State Reading} (SSR), which retains the VAE image space and SiT backbone while jointly denoising image latents, a global DINOv2~\cite{dinov2} CLS token, and a compact spatial patch state. Local states compressed by fixed PCA are fused with spatially corresponding image tokens at the input without increasing sequence length. Independent projections inject the current global state into every transformer block before attention, allowing deeper computations to use spatial information and directly access image-level context. During training, a frozen teacher supplies semantic targets, and the generator receives both joint-denoising and representation-alignment supervision. At inference, all three states evolve jointly from independent Gaussian noise, without a reference image or an additional semantic predictor.

Controlled ablations show that local-state modeling and direct global reading each improve the REG baseline, with their combination achieving the best FID at all three evaluated checkpoints. Reading at every layer also outperforms first-layer and alternating-layer reading. Semantic-trajectory interventions further change image layout while preserving the class label and image-noise path, demonstrating that the generated semantic branch participates in image formation.

We evaluate SiT-B/2, L/2, and XL/2 on ImageNet~\cite{imagenet} $256\times256$ and observe improved convergence across all three scales. On XL/2, SSR reaches or surpasses the reference FIDs of SiT, REPA, and REG using 100$\times$, 40$\times$, and 2$\times$ fewer training iterations, respectively. SSR achieves unguided FID 1.78 at 2M iterations and FID 1.34 with classifier-free guidance~\cite{cfg} alone. Combining it with autoguidance~\cite{autoguidance} yields FID 1.31 at 1M iterations, the best FID in our main comparison table. Our contributions are:
\begin{itemize}
\setlength{\itemsep}{2pt}
\setlength{\parskip}{0pt}
\item \textbf{Complementary global and local semantics.} We jointly model CLS and spatial patch states and use controlled ablations to examine the individual and combined benefits of local-state modeling and global reading.
\item \textbf{Per-layer semantic reading.} We fuse local states at corresponding input positions and let every block directly read the current global state, combining spatial correspondence with continuously accessible global context.
\item \textbf{The SSR joint-generation framework.} SSR improves convergence across three SiT scales with modest additional forward-pass overhead and substantially fewer training iterations to reach the reference quality levels above.
\end{itemize}

\section{Related Work}
\label{sec:related}

\paragraph{Diffusion transformers and semantic supervision.}
Latent diffusion~\cite{ldm} reduces image-generation costs by operating in a pretrained autoencoder space. DiT~\cite{dit} introduces a scalable transformer backbone, while SiT~\cite{sit} studies generative paths and sampling through stochastic interpolants. Despite these advances, learning semantic structure through denoising remains computationally demanding~\cite{mdt,maskdit,sd_dit}. REPA~\cite{repa} introduces semantic supervision by aligning intermediate generator features with clean pretrained representations. MaskAlign~\cite{maskalign} extends this approach through token mixing and subset alignment. These objectives supervise what intermediate features learn, but do not specify how each block accesses the current semantic state. SSR retains alignment while jointly generating global and local states, using spatially matched local input and direct per-layer global reading to make semantics available during both training and sampling.

\paragraph{Joint image--semantic modeling.}
Representation Entanglement for Generation (REG)~\cite{reg} jointly denoises image latents and a global CLS token, without an explicit spatially corresponding local state. ReDi~\cite{redi} jointly generates image latents and PCA-compressed patch features, without including a global CLS state in joint denoising. REGLUE~\cite{reglue} combines global CLS and nonlinearly compressed multi-layer spatial features, establishing a precedent for joint global--local modeling. SSR focuses on how these states are read: fixed-PCA patch states are fused with corresponding image tokens at the input, while independent projections give every block direct access to the current global state $c_t$, alongside the evolving CLS hidden token. This additional path complements input fusion and subsequent self-attention. The per-layer injection follows an access pattern related to Astra~\cite{astra}, applied here to internally generated semantics rather than external action conditions. SSR thus combines spatially organized local information with global context that remains directly accessible throughout network depth.

\begin{figure}[tbp]
\centering\includegraphics[width=.85\linewidth]{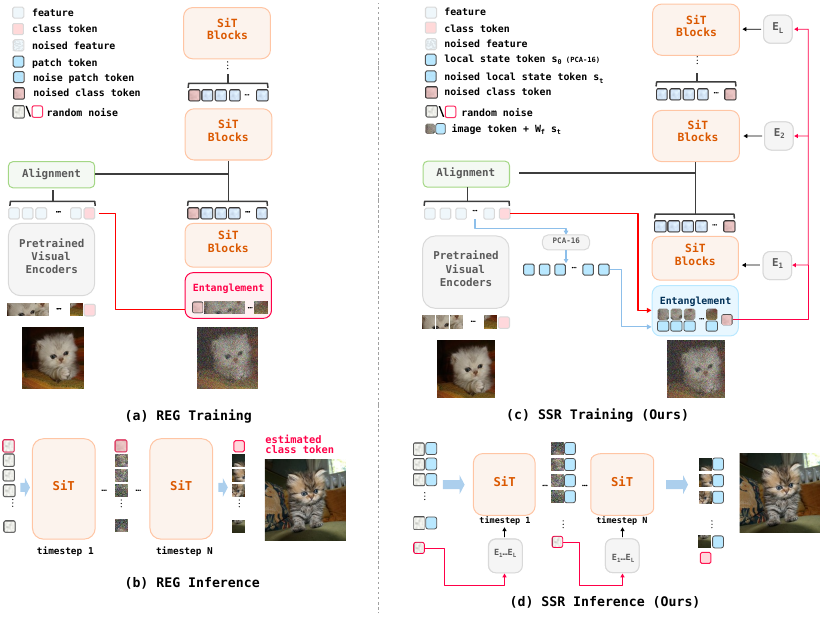}
\caption{\textbf{Comparison between REG and our Semantic State Reading (SSR).} (a) During training, REG jointly denoises image latents and a global CLS token from a pretrained visual encoder, while aligning intermediate features with clean teacher representations. (b) During inference, REG jointly generates image latents and global semantics from random noise. (c) SSR additionally models PCA-compressed spatial patch features as a local semantic state, which is fused with spatially corresponding image tokens at the input. Each transformer block also directly reads the current noisy CLS state through an independent projection $E_\ell$, providing global context throughout the network. (d) During inference, SSR jointly generates image, local and global states from random noise, retaining spatial input fusion and per-layer CLS injection at every sampling step. No reference image or visual encoder is required during sampling.}
\label{fig:overview}
\end{figure}

\section{Method}
\label{sec:method}

SSR jointly generates image latents, a global CLS state and a spatial patch state; the local state is fused with corresponding image tokens at the input, and the global state is read directly by every block (Fig.~\ref{fig:overview}).

\subsection{Preliminaries}
We build on SiT~\cite{sit} and the stochastic interpolant framework~\cite{stochastic}, using a linear path also employed in flow-matching formulations~\cite{flow,rectified}. Given a clean variable $u_0$ and independent Gaussian noise $\epsilon_u$, we use the linear interpolation path
\begin{equation}
    u_t=(1-t)u_0+t\epsilon_u,\qquad t\in[0,1].
    \label{eq:path}
\end{equation}
The model predicts the velocity $\epsilon_u-u_0$ from the current state, timestep and class label; sampling integrates the corresponding reverse-time SDE from Gaussian noise~\cite{sit,stochastic,score_sde}.

\subsection{Joint Global and Local Semantic Modeling}
A global CLS token provides image-level context without explicit positional correspondence, whereas spatial patch features supply position-specific semantic information. To use both, SSR jointly models image latents, a global CLS state and a local spatial state within a single diffusion process.

Given a clean image $I$, a frozen VAE encoder~\cite{vae,ldm} extracts image latents $x_0$, and a frozen DINOv2~\cite{dinov2} encoder extracts a global CLS token $c_0$ and spatial patch features $F_0$. Fixed PCA~\cite{pca,redi} with variance normalization gives a compact local state $s_0=\mathcal C_{\mathrm{PCA}}(F_0)$. For ImageNet $256\times256$, the image latents have size $32\times32\times4$, the CLS state has 768 dimensions, and the local state contains 256 vectors of dimension 16 on a $16\times16$ grid. PCA is fitted beforehand on teacher features from training images and remains frozen together with the teacher.

All three states follow Eq.~\eqref{eq:path} with a shared timestep $t$ and independent noise $\epsilon_x,\epsilon_c,\epsilon_s$. The network takes the joint state $(x_t,c_t,s_t)$ as input and predicts all three velocity components. The global state follows REG's joint denoising formulation~\cite{reg}, while the local state follows ReDi's PCA-based feature modeling~\cite{redi}. Their combination allows the model to generate both image-level context and spatial semantics during sampling, without externally supplied clean semantic conditions.

\subsection{Semantic State Reading}
\paragraph{Local spatial reading.}
Patch embedding converts image latents into a $16\times16$ token grid, matching the local state position by position. The input image tokens are
\begin{equation}
    H_{\mathrm{img}}^{(0)}
    =\operatorname{Embed}(x_t)+\phi_f(s_t)+\phi_r(\Pi s_t).
    \label{eq:local}
\end{equation}
Here, $\phi_f$ is a position-wise affine projection into the backbone hidden dimension. The operator $\Pi$ averages contiguous $4\times4$ regions and broadcasts their means back to the original grid; $\phi_r$ is a bias-free linear projection. Both projections are zero-initialized.

The local state is added to corresponding image tokens without increasing the number of sequence positions. The region branch reads means of the same local state and introduces no additional diffusion variable. The global CLS state is projected, normalized~\cite{layernorm} and concatenated with the image tokens:
\begin{equation}
    H_0=\bigl[\operatorname{LN}(\phi_c(c_t));H_{\mathrm{img}}^{(0)}\bigr]
    +E_{\mathrm{pos}},
    \label{eq:tokens}
\end{equation}
where $E_{\mathrm{pos}}$ denotes positional embeddings.

\paragraph{Per-layer global reading.}
When the global token is introduced only at the input, subsequent blocks access its semantics through progressively transformed hidden features. SSR retains this sequence structure while also letting every block directly read the CLS state $c_t$ of the current diffusion timestep.

For block $\ell$ with modulation condition $a=\operatorname{emb}(t)+\operatorname{emb}(y)$, an independent projection of the current CLS state is added after AdaLN~\cite{dit}:
\begin{equation}
\begin{aligned}
    U_\ell={}&\operatorname{AdaLN}_\ell(H_{\ell-1};a)\\
    &+\bigl[0;\operatorname{Broadcast}(E_\ell c_t+b_\ell)\bigr].
\end{aligned}
\label{eq:global}
\end{equation}
The additional projection is broadcast only to image tokens, not to the CLS position. Every block reads the same current state $c_t$, rather than the CLS hidden feature updated by the preceding block, using its own parameters $E_\ell,b_\ell$.

Attention and the residual update~\cite{transformer} then give
\begin{equation}
    H'_\ell=H_{\ell-1}+\gamma_\ell\odot
    P_\ell\!\left(\operatorname{Attn}_\ell(U_\ell)\right),
    \label{eq:attn}
\end{equation}
followed by the standard MLP branch. Here, $\gamma_\ell$ is SiT's conditional gate, and $P_\ell$ is an affine map applied to all tokens. The global projection is zero-initialized; the weight of $P_\ell$ is initialized to the identity and its bias to zero.

This placement follows Astra's per-layer injection design~\cite{astra}, but the injected signal is an internally generated semantic state. The local path preserves spatial correspondence, while the global path gives deeper blocks direct access to the current image-level context.

\subsection{Training and Inference}
\paragraph{Training objective.}
Three output heads predict velocities $v_x$, $v_c$ and $v_s$ for the image, global CLS and local states, conditioned on $(x_t,c_t,s_t,t,y)$. The image and local heads read the final image tokens, while the global head reads the CLS token. The joint prediction loss is
\begin{equation}
\begin{aligned}
    \mathcal L_{\mathrm{pred}}=\mathbb E\bigl[
    &\operatorname{MSE}(v_x,\epsilon_x-x_0)\\
    &+\beta\operatorname{MSE}(v_c,\epsilon_c-c_0)\\
    &+\mu\operatorname{MSE}(v_s,\epsilon_s-s_0)\bigr],
\end{aligned}
\label{eq:prediction}
\end{equation}
where each MSE is averaged over the elements of its corresponding prediction tensor.

We also retain REPA-style representation alignment~\cite{repa}. A projector $h$ maps hidden tokens at block $k$ to the teacher dimension, aligning them with clean teacher CLS and patch features:
\begin{equation}
    \mathcal L_{\mathrm{align}}
    =-\mathbb E\!\left[
    \frac{1}{N+1}\sum_{i=0}^{N}
    \cos\bigl(h(H_k)_i,T_i\bigr)
    \right].
    \label{eq:alignment}
\end{equation}
Here, $N=256$ and $T=[c_0;F_0]$ contains one global CLS token and 256 teacher patch features before PCA compression. Local velocity prediction targets the compressed state, whereas alignment uses the original teacher representations. The total objective is
\begin{equation}
    \mathcal L_{\mathrm{total}}
    =\mathcal L_{\mathrm{pred}}+\lambda\mathcal L_{\mathrm{align}}.
    \label{eq:loss}
\end{equation}
The default weights are $\beta=0.03$, $\mu=0.25$ and $\lambda=1$. Alignment is applied after block 4 for B/2 and block 8 for L/2 and XL/2. Training updates the generator and its reading, prediction and alignment modules, while the teacher and PCA remain frozen.

\paragraph{Inference.}
Image latents, global CLS and local states are initialized from independent Gaussian noise and jointly updated on the same Euler--Maruyama grid under a class label. Every sampling step uses the same reading paths as training. Sampling requires neither a teacher model nor a reference image or a separate semantic prediction network.

We train the unconditional branch with 10\% label dropout and optionally apply classifier-free guidance~\cite{cfg} separately to the three velocity components. The final step updates only the image latents, which are decoded by the VAE; the semantic states are then discarded. For deployment, the post-attention affine map can be merged into the attention output projection.

\section{Experiments}
\label{sec:experiments}
\subsection{Setup}
\paragraph{Training.}
We train class-conditional SiT-B/2, L/2 and XL/2 from scratch on ImageNet~\cite{imagenet} $256\times256$, using SD-VAE~\cite{ldm} latents of size $32\times32\times4$. Following REG~\cite{reg}, we use batch 256, AdamW~\cite{adamw} with learning rate $10^{-4}$, gradient clipping 1.0, EMA 0.9999 and 10\% label dropout. Alignment depth is 4 for B/2 and 8 for L/2 and XL/2. Each result uses one training seed. Same-platform main comparisons use REG with its published $\lambda=0.5$ and official-code REPA on L/2 and XL/2; controlled module ablations instead fix $\lambda=1$ for all four settings. Our REG runs agree with published results: at 200K, 6.54 versus 6.6 on L/2 and 4.86 versus 5.0 on XL/2; at 400K, 15.11 versus 15.2 on B/2.

\paragraph{Evaluation.}
We evaluate 50K samples using FID~\cite{fid}, spatial FID (sFID)~\cite{sfid}, Inception Score (IS)~\cite{is}, and precision/recall~\cite{precision}, with the ADM evaluation suite~\cite{adm}, EMA weights and a 250-step Euler--Maruyama SDE sampler. Results are unguided unless stated otherwise. Published comparisons are explicitly marked and use the values reported in their source papers. MaskAlign and REGLUE are external baselines with different implementations; neither has been rerun on our platform.

\begin{figure}[tbp]
\centering
\includegraphics[width=.82\linewidth]{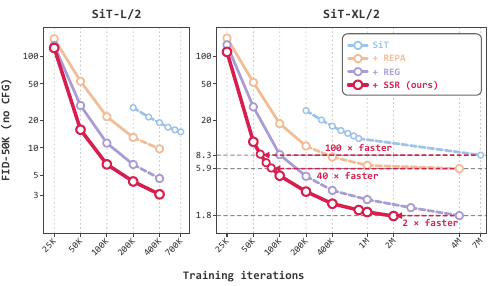}
\caption{\textbf{SSR accelerates convergence across model scales.} Unguided FID-50K on ImageNet $256\times256$ with SiT-L/2 and XL/2. On XL/2, SSR reaches or surpasses the reference FIDs of SiT, REPA, and REG using $100\times$, $40\times$, and $2\times$ fewer training iterations, respectively.}
\label{fig:curves}
\end{figure}

\begin{table}[tbp]
\centering
    \caption{\textbf{FID comparison across training iterations for diffusion transformers with semantic supervision.} All results are evaluated on ImageNet $256\times256$ without classifier-free guidance (CFG). $\dagger$ denotes results reported in the original papers.}
    \label{tab:main}
    \small
    \setlength{\tabcolsep}{3.5pt}
    {\begin{tabular*}{.48\linewidth}[t]{@{\extracolsep{\fill}}lrrr@{}}
      \toprule
      Method & \#Params & Iter. & FID$\downarrow$ \\
      \midrule
      SiT-B/2$^\dagger$~\cite{sit} & 130M & 400K & 33.0 \\
      + REPA$^\dagger$~\cite{repa} & 130M & 400K & 24.4 \\
      + ReDi$^\dagger$~\cite{redi,coredi} & 130M & 400K & 21.4 \\
      + REG~\cite{reg} & 132M & 100K & 35.18 \\
      + REG & 132M & 400K & 15.11 \\
      + MaskAlign$^\dagger$~\cite{maskalign} & 154M & 400K & 14.8 \\
      + REGLUE$^\dagger$~\cite{reglue} & 132M & 400K & 12.9 \\
      + \textbf{SSR (ours)} & 139M & 100K & 21.95 \\
      + \textbf{SSR (ours)} & 139M & 400K & \textbf{9.79} \\
      \midrule
      SiT-L/2$^\dagger$ & 458M & 400K & 18.8 \\
      + REPA & 458M & 100K & 21.88 \\
      + REPA$^\dagger$ & 458M & 400K & 9.7 \\
      + REG & 460M & 100K & 11.23 \\
      + REG$^\dagger$ & 460M & 400K & 4.6 \\
      + \textbf{SSR (ours)} & 479M & 100K & 6.56 \\
      + \textbf{SSR (ours)} & 479M & 400K & \textbf{3.08} \\
      \bottomrule
    \end{tabular*}\hspace{.035\linewidth}\begin{tabular*}{.48\linewidth}[t]{@{\extracolsep{\fill}}lrrr@{}}
      \toprule
      Method & \#Params & Iter. & FID$\downarrow$ \\
      \midrule
      SiT-XL/2$^\dagger$ & 675M & 400K & 17.2 \\
      SiT-XL/2$^\dagger$ & 675M & 7M & 8.3 \\
      + REPA & 675M & 100K & 18.38 \\
      + REPA$^\dagger$ & 675M & 4M & 5.9 \\
      + REG & 677M & 100K & 8.39 \\
      + REG$^\dagger$ & 677M & 400K & 3.4 \\
      + REG$^\dagger$ & 677M & 1M & 2.7 \\
      + REG$^\dagger$ & 677M & 2.4M & 2.2 \\
      + REG$^\dagger$ & 677M & 4M & 1.8 \\
      + REGLUE$^\dagger$ & 677M & 200K & 4.6 \\
      + REGLUE$^\dagger$ & 677M & 400K & 3.2 \\
      + REGLUE$^\dagger$ & 677M & 700K & 2.7 \\
      + REGLUE$^\dagger$ & 677M & 1M & 2.5 \\
      + MaskAlign$^\dagger$ & 732M & 200K & 4.0 \\
      + MaskAlign$^\dagger$ & 732M & 400K & 2.8 \\
      + MaskAlign$^\dagger$ & 732M & 1M & 2.4 \\
      + MaskAlign$^\dagger$ & 732M & 2.4M & 2.1 \\
      + \textbf{SSR (ours)} & 702M & 70K & 6.84 \\
      + \textbf{SSR (ours)} & 702M & 100K & 4.93 \\
      + \textbf{SSR (ours)} & 702M & 200K & 3.30 \\
      + \textbf{SSR (ours)} & 702M & 400K & 2.43 \\
      + \textbf{SSR (ours)} & 702M & 1M & 1.97 \\
      + \textbf{SSR (ours)} & 702M & 2M & \textbf{1.78} \\
      \bottomrule
    \end{tabular*}}
\end{table}

\subsection{Generation Quality and Convergence}

\paragraph{Accelerating training convergence.}
Table~\ref{tab:main} compares SSR with SiT, REPA, REG, REGLUE, and MaskAlign on ImageNet $256\times256$ without classifier-free guidance; Fig.~\ref{fig:curves} shows convergence across model scales. SSR consistently improves FID across model scales and evaluated common checkpoints. At 400K iterations, it achieves FID 9.79, 3.08 and 2.43 on B/2, L/2, and XL/2, respectively. On B/2, SSR improves over REG, MaskAlign and published REGLUE by 5.32, 5.01 and 3.11 FID points, respectively. On XL/2, its 200K, 400K and 1M FIDs of 3.30, 2.43 and 1.97 compare with REGLUE's published 4.6, 3.2 and 2.5 at the same iterations~\cite{reglue}. On XL/2, the gains emerge early: SSR reduces REG's FID from 28.03 to 11.61 at 50K and from 8.39 to 4.93 at 100K.

SSR also reaches the reference quality of longer-trained models with substantially fewer updates. SiT-XL/2 + SSR achieves FID 6.84 at 70K iterations, surpassing SiT at 7M (8.3)~\cite{sit}, and FID 4.93 at 100K, surpassing REPA at 4M (5.9)~\cite{repa}. These measured checkpoints use $1/100$ and $1/40$ of the respective reference iteration budgets. With longer training, SSR reaches FID 1.97 at 1M versus MaskAlign's 2.4 at that budget~\cite{maskalign}, and 1.78 at 2M versus REG's 1.8 at 4M~\cite{reg}. These are ratios of measured checkpoint budgets at reference quality levels, not measured wall-clock speedups or exact threshold-crossing times.

\paragraph{Generation quality.}
Beyond FID, SSR improves precision, IS, and sFID over REG at 200K on XL/2: precision increases from 0.751 to 0.786, IS increases from 161.7 to 185.8, and sFID decreases from 4.99 to 4.66. With longer training, SSR reaches IS 228.4 and sFID 4.44 at 2M. Figure~\ref{fig:compare} compares REPA, REG, and SSR at 50K, 100K, and 200K using shared labels and image noise.

\begin{table}[tbp]
\centering
\caption{\textbf{Guided generation on ImageNet $256\times256$.} Published baselines use their reported guidance settings. SSR (CFG) uses scale 2.3 for $t\leq0.85$; SSR (hybrid) adds autoguidance at extra sampling cost. Arrows indicate the preferred direction.}
\label{tab:guided}
    \small
    \setlength{\tabcolsep}{3.5pt}
    {\begin{tabular*}{.96\linewidth}{@{\extracolsep{\fill}}lrrrrrr@{}}
      \toprule
      Method & Epochs & FID$\downarrow$ & sFID$\downarrow$ & IS$\uparrow$ & Pre.$\uparrow$ & Rec.$\uparrow$ \\
      \midrule
      DiT-XL/2~\cite{dit} & 1400 & 2.27 & 4.60 & 278.2 & \textbf{0.83} & 0.57 \\
      SiT-XL/2~\cite{sit} & 1400 & 2.06 & 4.50 & 270.3 & 0.82 & 0.59 \\
      + REPA~\cite{repa} & 800 & 1.42 & 4.70 & 305.7 & 0.80 & 0.65 \\
      + REG~\cite{reg} & 80 & 1.86 & 4.49 & \textbf{321.4} & 0.76 & 0.63 \\
      + REG & 160 & 1.59 & 4.36 & 304.6 & 0.77 & 0.65 \\
      + REG & 480 & 1.40 & 4.24 & 296.9 & 0.77 & 0.66 \\
      + REG & 800 & 1.36 & 4.25 & 299.4 & 0.77 & 0.66 \\
      + MaskAlign~\cite{maskalign} & 80 & 1.82 & 4.48 & 310.0 & 0.81 & 0.63 \\
      + MaskAlign & 160 & 1.56 & 4.37 & 304.1 & 0.79 & 0.65 \\
      + MaskAlign & 800 & 1.35 & 4.31 & 312.9 & 0.78 & \textbf{0.67} \\
      + REGLUE~\cite{reglue} & 80 & 1.59 & 4.22 & 282.9 & 0.78 & 0.64 \\
      + REGLUE & 160 & 1.46 & 4.23 & 293.9 & 0.78 & 0.65 \\
      \midrule
      + \textbf{SSR (ours, CFG)} & 80 & 1.51 & 4.31 & 289.4 & 0.80 & 0.62 \\
      + \textbf{SSR (ours, CFG)} & 160 & 1.42 & 4.20 & 297.0 & 0.80 & 0.64 \\
      + \textbf{SSR (ours, CFG)} & 200 & 1.39 & \textbf{4.12} & 300.1 & 0.80 & 0.63 \\
      + \textbf{SSR (ours, CFG)} & 400 & 1.34 & 4.23 & 303.6 & 0.80 & 0.65 \\
      \midrule
      + \textbf{SSR (ours, hybrid)} & 80 & 1.47 & 4.75 & 266.5 & 0.79 & 0.64 \\
      + \textbf{SSR (ours, hybrid)} & 160 & 1.33 & 4.80 & 277.5 & 0.79 & 0.65 \\
      + \textbf{SSR (ours, hybrid)} & 200 & \textbf{1.31} & 4.73 & 279.7 & 0.79 & 0.65 \\
      \bottomrule
    \end{tabular*}}
\end{table}

\paragraph{Comparison under guided sampling.}
Table~\ref{tab:guided} compares guided generation across training budgets. On XL/2, CFG~\cite{cfg,interval} 2.3 applied to all three states for $t\leq0.85$ yields FID 1.51, 1.42, 1.39 and 1.34 at 80, 160, 200 and 400 epochs. At 80/160 epochs, these results improve on the published REG (1.86/1.59)~\cite{reg}, MaskAlign (1.82/1.56)~\cite{maskalign} and REGLUE (1.59/1.46)~\cite{reglue} values. The gaps to REGLUE are 0.08/0.04 FID; repeated-seed uncertainty is not available. This is a comparison of reported protocols: REGLUE uses CFG 2.3 for $t\leq0.90$~\cite{reglue}. At 400 epochs, SSR is numerically below REG's 1.36 and MaskAlign's 1.35 at 800 epochs. Pure CFG attains sFID 4.12 at 200 epochs; IS rises from 289.4 to 303.6 across the four checkpoints, while precision remains 0.80. A separate B/2 test at 400K supports guiding the local state: guiding all three states gives FID 4.44 versus 5.79 without local-state guidance, at the same CFG scale and interval.

We further combine CFG with autoguidance~\cite{autoguidance}:
\begin{equation}
 d=d_{\mathrm{CFG}}+(w_{\mathrm{ag}}-1)
 (d_{\mathrm{cond}}-d_{\mathrm{weak}}),
\end{equation}
using SiT-L/2 + SSR at 100K as the weak model. With CFG 1.8 and $w_{\mathrm{ag}}=1.5$, this hybrid configuration achieves FID 1.47, 1.33 and 1.31 at 80, 160 and 200 epochs, respectively: 0.04, 0.09 and 0.08 below the pure-CFG configuration. Because the CFG scale also changes, these gaps do not isolate the effect of autoguidance alone. The measured sampling time is $1.39\times$ that of CFG alone. The 1.31 FID is the lowest among the listed VAE-based comparisons and requires an additional weak model; it is not a claim of state of the art across all image representations. At matched epochs, pure CFG has lower sFID and higher IS and precision than hybrid guidance; the two schemes offer different metric and compute trade-offs.

\begin{figure}[tbp]
\centering
\includegraphics[width=.82\linewidth]{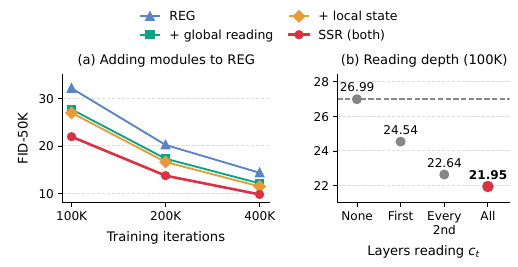}
\caption{\textbf{Global--local semantic modeling and global reading.} Unguided FID-50K on SiT-B/2 with $\lambda=1$. (a) Adding either module improves REG; combining them yields the lowest FID at all three checkpoints. (b) With the local state present, reading the global state at every layer achieves the best FID at 100K iterations.}
\label{fig:ablation}
\end{figure}

\begin{table}[tbp]\centering\small
\caption{\textbf{Ablations of semantic representations, reading designs, and training configurations.} Unguided results at 200K iterations on SiT-B/2 with a shared seed (FID-50K). The reference configuration is SSR with DINOv2-B, PCA-16, $k=4$, $\beta=0.03$, $\lambda=1$, $\mu=0.25$, and per-layer $E_\ell,P_\ell$. Each group varies the indicated factor; the $\mu$ sweep fixes $\lambda=0.5$. REG + ReDi~\cite{reg,redi} omits global reading and the regional-mean branch, which the local-only variant in Fig.~\ref{fig:ablation}(a) retains; \emph{Pre-merged} merges $P_\ell$ before training; \emph{Read CLS of $H_{\ell-1}$} lets $E_\ell$ read the normalized CLS row of the preceding hidden state instead of $c_t$. Bold denotes the best value in each column.}
\label{tab:ablation}
\renewcommand{\arraystretch}{1.04}
\setlength{\tabcolsep}{2.2pt}
\begin{tabular*}{\linewidth}{@{\extracolsep{\fill}}llrrrrr@{}}\toprule
Ablation & Setting & FID$\downarrow$ & sFID$\downarrow$ & IS$\uparrow$ & Pre.$\uparrow$ & Rec.$\uparrow$\\\midrule
\multicolumn{2}{@{}l}{SSR (default)} & \textbf{13.74} & 6.74 & \textbf{99.3} & \textbf{0.679} & 0.606\\\midrule
Baselines & REG & 20.22 & 6.98 & 78.2 & 0.614 & \textbf{0.624}\\
 & \shortstack[l]{REG + ReDi,\\PCA-8} & 16.89 & 6.74 & 90.6 & 0.658 & 0.616\\
 & \shortstack[l]{REG + ReDi,\\PCA-16} & 16.84 & 6.80 & 93.3 & 0.661 & 0.617\\\midrule
PCA channels & 8 & 15.05 & 6.93 & 96.2 & 0.665 & 0.599\\
 & 12 & 13.94 & 6.82 & 98.9 & 0.673 & 0.607\\
 & 16 (default) & \textbf{13.74} & 6.74 & \textbf{99.3} & \textbf{0.679} & 0.606\\\midrule
Global read & Shared $E,P$ & 14.61 & 6.93 & 95.0 & 0.669 & 0.613\\
 & Shared $E$, no $P$ & 14.35 & 6.90 & 95.8 & 0.668 & 0.607\\
 & $P_\ell$ pre-merged & 14.16 & 6.88 & 96.1 & 0.673 & 0.609\\
 & \shortstack[l]{Read CLS of\\$H_{\ell-1}$} & 13.80 & \textbf{6.59} & 97.6 & 0.669 & 0.611\\\midrule
Target & DINOv2-L & 15.16 & 7.29 & 96.9 & 0.666 & 0.595\\
 & CLIP-L & 21.18 & 7.18 & 71.3 & 0.627 & 0.606\\\midrule
Depth $k$ & 2 & 14.24 & 6.77 & 94.9 & 0.675 & 0.600\\
 & 6 & 15.18 & 7.00 & 93.4 & 0.669 & 0.595\\\midrule
$\beta$ & 0.05 & 14.21 & 6.91 & 97.4 & 0.669 & 0.610\\
 & 0.10 & 15.42 & 7.32 & 94.9 & 0.659 & 0.601\\\midrule
$\lambda$ & 0 & 20.36 & 6.69 & 71.2 & 0.647 & 0.593\\
 & 0.25 & 15.52 & 6.91 & 91.6 & 0.669 & 0.601\\
 & 0.5 & 14.29 & 6.85 & 95.3 & 0.674 & 0.602\\
 & 2 & 13.76 & 6.85 & 99.0 & 0.675 & 0.610\\\midrule
$\mu$ ($\lambda=0.5$) & 0 & 168.25 & 77.34 & 6.4 & 0.103 & 0.103\\
 & 0.125 & 14.85 & 6.84 & 94.4 & 0.670 & 0.597\\
 & 0.25 & 14.29 & 6.85 & 95.3 & 0.674 & 0.602\\
 & 0.5 & 14.75 & 7.04 & 94.2 & 0.670 & 0.596\\
 & 1 & 16.35 & 7.25 & 88.7 & 0.662 & 0.595\\
 & 2 & 23.66 & 8.72 & 70.0 & 0.609 & 0.588\\\bottomrule
\end{tabular*}\end{table}

\subsection{Ablation Studies}
\label{sec:exp_ablation}
We conduct SiT-B/2 ablations with a shared training seed. Figure~\ref{fig:ablation} evaluates module combinations and reading depth; Table~\ref{tab:ablation} reports five metrics at 200K. Defaults are DINOv2-B, PCA-16, $k=4$, $\beta=0.03$, $\lambda=1$ and $\mu=0.25$; the $\mu$ sweep uses $\lambda=0.5$. Default SSR has the best FID, IS and precision in Table~\ref{tab:ablation}, while hidden-state CLS reading has the lowest sFID and REG the highest recall.

\paragraph{Local-state modeling and global reading.}
Figure~\ref{fig:ablation}(a) isolates the two components at $\lambda=1$. At 100K, REG, global reading alone, local-state modeling alone and full SSR yield FID 32.16, 27.72, 26.99 and 21.95. Their corresponding 200K values are 20.22, 17.32, 16.62 and 13.74; at 400K they are 14.35, 12.12, 11.46 and 9.79. Local-state modeling has the larger standalone gain, but global reading improves it further by 5.04/2.88/1.67 FID. The joint gain minus the sum of the standalone gains is $+0.60/-0.02/-0.56$ at 100K/200K/400K. Thus both paths are useful, while their interaction is not uniformly additive or super-additive. These single-seed results do not establish interaction significance.

\paragraph{Reading across network depth.}
With the local state present, no direct reading, first-layer reading, alternating-layer reading and all-layer reading yield FID 26.99, 24.54, 22.64 and 21.95 at 100K (Fig.~\ref{fig:ablation}(b)). Sparse reading also helps; all-layer reading performs best among these choices. The separate source comparison at 200K is less decisive: reading $c_t$ gives FID 13.74 versus 13.80 for the normalized CLS row of $H_{\ell-1}$, but the hidden-state variant has better sFID (6.59 versus 6.74) and recall (0.611 versus 0.606).

\paragraph{Spatial-state dimension.}
PCA dimensions 8, 12 and 16 yield FID 15.05, 13.94 and 13.74 at 200K. PCA-16 also has the best IS and precision in this sweep; PCA-12 has slightly higher recall (0.607 versus 0.606). In the separate REG + ReDi baseline, increasing PCA-8 to PCA-16 changes FID only from 16.89 to 16.84, while sFID rises from 6.74 to 6.80. Both REG + ReDi and the local-only variant omit direct global reading, but the latter retains $\phi_r(\Pi s_t)$ and reaches 16.62 at 200K (Fig.~\ref{fig:ablation}(a)).

\paragraph{Per-layer parameterization.}
Independent $E_\ell,P_\ell$ give FID 13.74 at 200K, compared with 14.61 for shared $E,P$, 14.35 for shared $E$ without $P$, and 14.16 when $P_\ell$ is merged before training. This favors independent projections and a separate training parameterization for $P$ in FID. Merging $P$ after training is an algebraic reparameterization of the trained attention output, rather than the pre-merged training variant.

\paragraph{Teacher representations and alignment depth.}
DINOv2-B gives FID 13.74 at 200K, versus 15.16 for DINOv2-L and 21.18 for CLIP-L~\cite{clip}. Alignment at layer 4 gives 13.74, versus 14.24 and 15.18 at layers 2 and 6. These results support the default teacher and depth for B/2 under this training setup, not a general ranking of representation quality.

\paragraph{Loss weights.}
At 200K, $\beta=0.03$ gives FID 13.74, versus 14.21/15.42 for $0.05/0.10$. Raising $\lambda$ from 0 to 1 lowers FID from 20.36 to 13.74; $\lambda=2$ gives a similar 13.76. With $\lambda=0.5$ fixed, $\mu=0.25$ gives the lowest FID (14.29), versus 14.85/14.75 for $0.125/0.5$, and 16.35/23.66 for $1/2$. Removing local prediction ($\mu=0$) gives FID 168.25, sFID 77.34 and IS 6.4. This variant retains the joint-state architecture and sampling procedure: it is not equivalent to removing the local module. The severe degradation therefore supports training the sampled local state, without proving a unique mechanism or an optimal loss weight.

\begin{figure}[tbp]
\centering
\newcommand{\cimg}[1]{\includegraphics[width=.194\linewidth]{compare/#1.jpg}}
\newcommand{\rlab}[1]{\rotatebox{90}{\makebox[.194\linewidth]{\footnotesize #1}}}
\setlength{\tabcolsep}{0.5pt}\renewcommand{\arraystretch}{0.5}
\begin{tabular}{@{}c@{\hspace{2pt}}ccc@{}}
& {\footnotesize 50K} & {\footnotesize 100K} & {\footnotesize 200K}\\[2pt]
\rlab{REPA} & \cimg{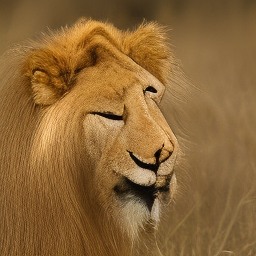} & \cimg{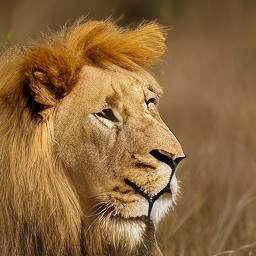} & \cimg{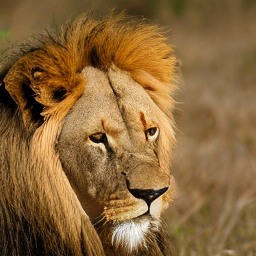}\\
\rlab{REG} & \cimg{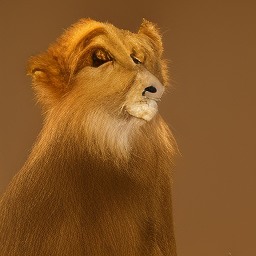} & \cimg{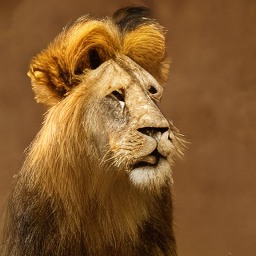} & \cimg{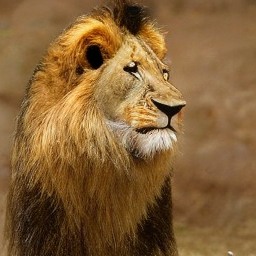}\\
\rlab{SSR} & \cimg{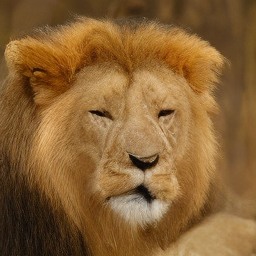} & \cimg{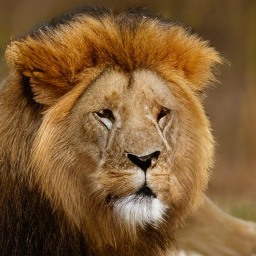} & \cimg{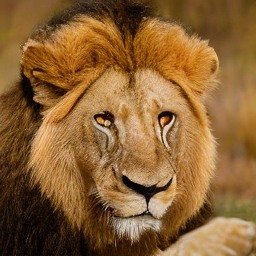}\\
\end{tabular}
\caption{\textbf{Samples across training iterations.} Same-platform SiT-XL/2 samples with REPA, REG and SSR at 50K, 100K and 200K iterations. All use the same label and the same initial and per-step image noise; CFG 4.0 on every generated variable, 250-step SDE, EMA. The class (lion) was selected for display from 12 classes fixed before sampling.}
\label{fig:compare}
\end{figure}

\begin{figure}[tbp]
\centering\includegraphics[width=\linewidth,trim=249 0 0 0,clip]{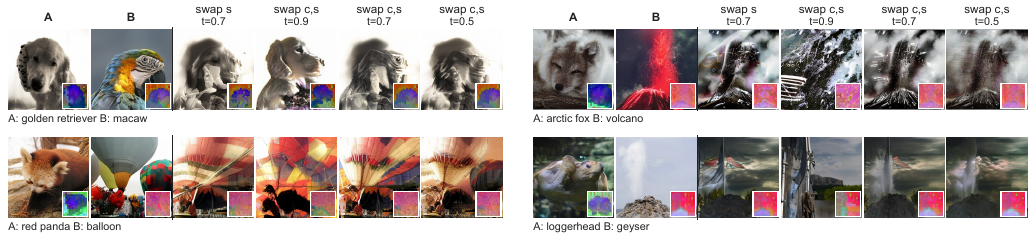}
\caption{\textbf{Semantic-trajectory swaps.} SiT-XL/2 + SSR at 2M, no guidance, seed 0. Two of the four predetermined class pairs are shown. Except column B, each column keeps A's label, image latent and image noise. At the indicated time, generated $s_t$, or both $c_t,s_t$, are replaced with B's and continue with B's semantic noise. Insets show the final local state using its first three directions as RGB.}
\label{fig:swap}
\end{figure}

\subsection{How the Generated States Affect the Image}
\label{sec:state_effects}
\paragraph{Intervening on generated semantics.}
A central question in joint image--semantic generation is whether the generated semantic states actively guide image formation or merely serve as auxiliary prediction targets. We address this question by intervening on the semantic branch during sampling. Using SiT-XL/2 + SSR trained for 2M iterations, we sample paired trajectories A and B without guidance with a 250-step SDE sampler. At time $t$, we replace A's local state $s_t$, or both its global and local states $(c_t,s_t)$, with B's states at the same time and continue with B's semantic noise. A's class label, image latent at the intervention, and subsequent image noise remain unchanged.

\paragraph{Layout changes under a fixed class label.}
Figure~\ref{fig:swap} shows two predetermined class pairs: arctic fox--volcano and loggerhead turtle--geyser. Swapping only $s_t$ at $t=0.7$ introduces a volcano-like cone or a geyser-like column despite retaining A's class label. Joint swaps also change content and structure, with outcomes depending on intervention time. Insets show the final local states through their first three PCA directions. These interventions change the image while its class condition and image-noise realization are held fixed. They also replace subsequent semantic noise, so they test the semantic trajectory rather than isolate a single state value.

Across four seeds and four class pairs, swapping $s_t$ at $t=0.9$ or $0.7$, or $c_t$ alone at $t=0.9$, displaces the final image latent from the unmodified A by 0.55, 0.43, and 0.36 of the original A--B distance, respectively. Together, these observations show sensitivity to the generated semantic trajectory, including layout changes beyond the class label. They do not establish that global and local states independently control content and layout.

\begin{table}[tbp]\centering\small
\caption{\textbf{Computational cost and generation performance on SiT-XL/2.} FID and IS are evaluated at 200K iterations. Parameters and FLOPs exclude the alignment projector. FLOPs are measured per forward pass at batch size 1; latency is the median of 30 FP32 forward passes at batch size 125 on the same device.}
\label{tab:cost}
\setlength{\tabcolsep}{3pt}
\begin{tabular*}{.94\linewidth}{@{\extracolsep{\fill}}lrrrrr@{}}\toprule
Method&Params (M)&GFLOPs&Latency (s)&FID&IS\\\midrule
REPA&\textbf{675.11}&\textbf{237.2}&\textbf{0.588}&10.44&105.7\\
REG&676.89&238.2&0.663&4.86&161.7\\
SSR&701.75&238.3&0.681&\textbf{3.30}&\textbf{185.8}\\\bottomrule
\end{tabular*}\end{table}

\subsection{Computational Cost}
Table~\ref{tab:cost} compares same-platform SiT-XL/2 inference costs. At 200K, SSR reduces REG's FID from 4.86 to 3.30 (32.1\%) and increases IS from 161.7 to 185.8 (14.9\%). It adds less than 0.1\% FLOPs, 2.7\% measured latency and 3.7\% parameters overall. Each global projection $E_\ell$ is computed once per image and layer, then broadcast across tokens. Thus the reading design improves generation quality at a small arithmetic and measured latency overhead.

\subsection{Limitations}
Due to the substantial computational resources required for training, all training comparisons use a single shared random seed. Although the state-intervention study uses four random seeds, these vary sampling randomness rather than training randomness. 

Our experiments are limited to ImageNet $256\times256$, and semantic swaps do not establish fully disentangled global and local control. Future work could evaluate multiple training seeds and extend the experiments to additional datasets and resolutions to further assess robustness and generalization.

\section{Conclusion}
We presented SSR to improve diffusion-transformer training through joint global--local semantic modeling and per-layer global reading. Our experiments show that adding spatial patch states to global semantic modeling improves generation, and directly exposing each block to the current global state provides further gains. Together, these findings support combining complementary semantic representations with explicit access throughout network depth. Semantic-trajectory swaps further reveal that the generated states influence image layout even when the class label and image-noise path remain fixed, connecting the jointly learned semantics to the generation process itself. This design improves convergence across three SiT scales with modest inference overhead. On SiT-XL/2, SSR surpasses REPA's reported reference FID with $40\times$ fewer training iterations and matches REG's reference quality with half the iterations. SSR thus demonstrates the value of coupling global--local semantic modeling with sustained semantic access for efficient image generation.

{\fontsize{9.7}{12}\selectfont
}
\clearpage
\endgroup
\newgeometry{twocolumn=false,left=.95in,right=.95in,top=.76in,bottom=.80in,headheight=12pt,headsep=25pt,footskip=30pt}
\onecolumn
\pagenumbering{arabic}
\renewcommand{\thepage}{S\arabic{page}}
\pagestyle{plain}
\raggedbottom
\let\section\AppendixSection
\let\subsection\AppendixSubsection
\let\subsubsection\AppendixSubsubsection
\renewcommand{\ttdefault}{pcr}
\hypersetup{hidelinks}
\setcounter{figure}{0}
\setcounter{table}{0}
\setcounter{equation}{0}
\renewcommand{\theHfigure}{supp.\arabic{figure}}
\renewcommand{\theHtable}{supp.\arabic{table}}
\renewcommand{\theHequation}{supp.\arabic{equation}}
\pdfbookmark[0]{Supplementary Material}{supplementary-material}
\captionsetup{font=small,labelfont=bf,labelsep=period,justification=raggedright,singlelinecheck=false,hypcap=false}
\captionsetup[table]{position=above,skip=5pt}
\captionsetup[figure]{position=below,skip=5pt}
\setlength{\parindent}{0pt}
\setlength{\parskip}{5pt plus .5pt minus .5pt}
\setlength{\abovedisplayskip}{7pt}
\setlength{\belowdisplayskip}{7pt}
\setlength{\bibsep}{3pt plus 1pt}
\renewcommand{\thefigure}{S\arabic{figure}}
\renewcommand{\thetable}{S\arabic{table}}
\renewcommand{\theequation}{S\arabic{equation}}
\newcommand{\Fifty}{FID-50K}
\newcommand{\field}[1]{\textbf{#1.}\enspace}
\normalsize
\begin{center}
{\Large\bfseries Semantic State Reading for Joint Image--Semantic\\[2pt] Diffusion Transformers\par}
\vspace{5pt}
{\large Supplementary Material\par}
\end{center}

\appendix
Sections~\ref{sec:implementation} and~\ref{sec:cost} describe the model and quantify its training and inference costs. Section~\ref{sec:modules} compares the reading paths and guidance settings. Section~\ref{sec:mechanism} tests which information the semantic states recover and how changing them affects an image. Section~\ref{sec:samples} shows paired samples, including complete sets of draws.
\section{Architecture and implementation}
\label{sec:implementation}
\field{Three states on one image grid}The frozen DINOv2-B/14 encoder~\cite{supp-dinov2} receives an ImageNet-normalized RGB crop resized to \(224\times224\) by bicubic interpolation. After the encoder's final LayerNorm, row zero gives the 768-dimensional \(c_0\) at its native scale, without further centering, normalization or rescaling. The remaining 256 rows give \(F_{0,i}\) in row-major order on a \(16\times16\) grid. A frozen SD-VAE~\cite{supp-ldm} encodes the \(256\times256\) image into a \(32\times32\times4\) latent \(x_0\). Each \(2\times2\) latent patch gives one SiT image token~\cite{supp-sit} at the corresponding grid location.

\field{Fitting the local state}One PCA basis is fitted across all positions of 20,000 ImageNet training images (5,120,000 patch vectors). With mean \(m\), covariance eigenvectors \(q_j\) and eigenvalues \(\nu_j\) in descending order, the whitened coordinates of patch \(i\) are
\begin{equation}
 s_{0,i,j}=q_j^{\mathsf T}(F_{0,i}-m)/\sqrt{\nu_j},\qquad j=1,\ldots,16 .
 \label{eq:pca}
\end{equation}
In an FP32 refit of the PCA basis in Eq.~\eqref{eq:pca} on the same images, the leading 8, 12 and 16 directions explain 13.5\%, 16.5\% and 19.0\% of the total variance before whitening. In the single-seed B/2 comparison at 200K updates, the corresponding unguided FIDs are 15.05, 13.94 and 13.74 (Table~3 of the main paper). The 16-channel choice gives the lowest measured FID in this sweep.

\begin{center}\begin{minipage}{\linewidth}
\captionof{table}{\textbf{Modules of SSR.} \(d\) is the backbone width. Every block has its own \(E_\ell\) and \(P_\ell\). An affine map includes a bias unless specified otherwise.}
\label{tab:states}\centering\small
\renewcommand{\arraystretch}{1.14}
\begin{tabularx}{\linewidth}{@{}p{.21\linewidth}X@{}}
\toprule Module & Operation and initialization\\\midrule
Local input \(\phi_f,\phi_r\) & Add \(\phi_f(s_{t,i})\) and \(\phi_r(\bar s_{t,R(i)})\) to image token \(i\), with \(\bar s_{t,R(i)}\) the mean of its contiguous \(4\times4\) region. Both maps are \(16\to d\) and start at zero. Only \(\phi_f\) has a bias.\\
Input CLS \(\phi_c\) & Affine \(768\to d\) with Xavier-uniform weights and zero bias, followed by learned LayerNorm (\(\epsilon=10^{-6}\)). Prepend this row to the 256 image rows. Add a fixed zero position vector to the CLS row and fixed 2D sine--cosine embeddings to the image rows.\\
Global read \(E_\ell\) & Zero-initialized affine \(768\to d\). Broadcast \(E_\ell(c_t)\) to the conditioned, normalized image rows immediately before self-attention. The CLS row receives no broadcast.\\
Attention output \(P_\ell\) & Affine \(d\to d\) on the attention output of all 257 rows, initialized to the identity with zero bias, before the gated residual update.\\
Velocity heads & A final conditioned LayerNorm feeds three zero-initialized affine heads. Image rows give \(v_x\) (\(d\to16\), reshaped into \(2\times2\times4\) patches) and \(v_s\) (\(d\to16\)). The CLS row gives \(v_c\) (\(d\to768\)).\\
Alignment head \(h\) & Training-only MLP \(d\to2048\to2048\to768\), with SiLU after each hidden affine layer, Xavier-uniform weights and zero biases. All 257 rows are aligned to the uncompressed teacher features.\\
\bottomrule
\end{tabularx}\end{minipage}\end{center}
All three backbones use the modules in Table~\ref{tab:states}. B/2, L/2 and XL/2 use 12, 24 and 28 blocks, widths 768, 1024 and 1152, and 12, 16 and 16 attention heads, respectively. The alignment head reads block 4 in B/2 and block 8 in L/2 and XL/2.

\clearpage
\subsection{Training and sampling}
\field{One training update}We draw \(x_0=0.18215\,(m_x+\sigma_x\odot\eta)\), with \(\eta\sim\mathcal N(0,I)\), from the cached SD-VAE posterior mean and standard deviation in \texttt{SwayStar123/repa-imagenet-256} (revision \texttt{21fc92f23f5f}). Draw one \(t\sim\mathcal U(0,1)\) per image independently of the posterior sample. For each \(u\in\{x,c,s\}\), draw independent standard Gaussian noise \(\epsilon_u\) and set \(u_t=(1-t)u_0+t\epsilon_u\). The three heads predict \(\epsilon_u-u_0\) with mean squared error weights \(1,0.03,0.25\) for \(x,c,s\). Add the negative cosine alignment loss~\cite{supp-repa} with weight 1. Drop the class label with probability 0.1. AdamW uses a global batch of 256, constant learning rate \(10^{-4}\), moments \((0.9,0.999)\), \(\epsilon=10^{-8}\), zero weight decay and gradient-norm clipping at 1. The teacher and student use FP16 autocast. Student parameters, AdamW state, PCA projection and loss reductions use FP32. We use dynamic loss scaling and unscale gradients before clipping. Only successful updates advance the step count and the EMA (decay 0.9999).

\field{One network evaluation}At inference the three initial states are independent standard Gaussians. Each block reads the same current \(c_t\) through its own \(E_\ell\) (Table~\ref{tab:states}). The hidden CLS row evolves through attention and the MLP. The updated diffusion state supplies \(c_t\) at the next solver step. The final LayerNorm has no learned affine parameters of its own. A SiLU and a \(d\to2d\) affine map of the timestep and label conditioning supply its shift and scale.

\field{Updating the three states}For a velocity \(v_u\) at positive \(t\), the linear interpolation path gives the estimated score \(\widehat q_u\), diffusion variance \(g^2\) and reverse-time drift \(d_u\) below.
\begin{align}
 \widehat q_u &= -\big[u_t+(1-t)v_u\big]/t,\qquad
 g^2(t)=2t,\qquad d_u=v_u-\tfrac12g^2(t)\widehat q_u ,
 \label{eq:score}\\
 u_{t+\Delta t} &= u_t+d_u\,\Delta t+\sqrt{2t\,|\Delta t|}\,\xi_u,\qquad \Delta t<0 .
 \label{eq:em}
\end{align}
The Euler--Maruyama grid has 250 positive evaluation times spaced uniformly from 1 to 0.04. For each of the first 249 steps we draw fresh, independent standard Gaussian increments \(\xi_u\). From \(t=0.04\) to zero, we take a deterministic final drift step on \(x\), divide the resulting latent by \(0.18215\) and decode it with \texttt{stabilityai/sd-vae-ft-ema}. For guided sampling the drift in Eq.~\eqref{eq:em} is combined separately for each stream as specified in Section~\ref{sec:evaluation}.

\subsection{Parameter accounting}
\begin{center}\begin{minipage}{\linewidth}
\captionof{table}{\textbf{Student parameters in millions.} Trainable counts include the alignment head and exclude fixed position embeddings. Generator counts include fixed positions and exclude the alignment head. Frozen encoders and the EMA copy are excluded throughout.}
\label{tab:params}\centering\small
\begin{tabular*}{\linewidth}{@{\extracolsep{\fill}}lrrrrr@{}}
\toprule &\multicolumn{2}{c}{Trainable student}&Unmerged generator&\multicolumn{2}{c}{Inference generator}\\
\cmidrule(lr){2-3}\cmidrule(lr){4-4}\cmidrule(lr){5-6}
Backbone & REG & SSR & SSR & REG & SSR (merged)\\\midrule
SiT-B/2 & 138.83 & 153.04 & 145.90 & 131.68 & 138.81\\
SiT-L/2 & 467.27 & 511.41 & 503.80 & 459.66 & 478.61\\
SiT-XL/2 & 684.72 & 746.77 & 738.94 & 676.89 & 701.75\\\bottomrule
\end{tabular*}\end{minipage}\end{center}
The XL/2 alignment head has 8.13M trainable parameters. Each of the 28 blocks has a post-attention map with \(1152(1152+1)\) parameters, for 37.19M in total before merging (Table~\ref{tab:params}). Its global read has \(1152(768+1)\) parameters. Across blocks, these reads retain 24.80M parameters at inference.

\clearpage
\section{Training and inference cost}
\label{sec:cost}
\field{Observed training cost}At its 200K checkpoint, SSR reaches unguided FID 13.74 in 64.11 logical NPU-hours. REG reaches 14.35 in 113.93 NPU-hours at 400K (Table~\ref{tab:observedcost}). SSR uses 43.7\% fewer training-loop NPU-hours for these two measured checkpoints. At equal 400K budgets, SSR uses 8.95\% more NPU-hours and lowers FID from 14.35 to 9.79.
\begin{center}
\begin{minipage}{\linewidth}
\captionof{table}{\textbf{Observed training-loop cost on Ascend 910C.} Cumulative logical NPU-hours and unguided FID-50K for SiT-B/2. Both runs use four logical devices, global batch size 256 and alignment weight \(\lambda=1\), as in Fig.~4 of the main paper. The REG baseline in Table~1 of the main paper uses \(\lambda=0.5\).}
\label{tab:observedcost}
\centering\small
\begin{tabular*}{.88\linewidth}{@{\extracolsep{\fill}}lrrrr@{}}
\toprule & \multicolumn{2}{c}{REG} & \multicolumn{2}{c}{SSR}\\
\cmidrule(lr){2-3}\cmidrule(lr){4-5}
Updates & FID & NPU-hours & FID & NPU-hours\\
\midrule
100K & 32.16 & 29.23 & 21.95 & 30.59 \\
200K & 20.22 & 57.49 & 13.74 & 64.11 \\
400K & 14.35 & 113.93 & 9.79 & 124.13 \\
\bottomrule
\end{tabular*}
\end{minipage}
\end{center}

Each B/2 run used four logical Ascend 910C devices on its own host. We sum training-loop elapsed time across three consecutive segments and multiply by four. The logged intervals include skipped mixed-precision update attempts. Model initialization, offline data preparation, gaps between launches, final checkpoint serialization and evaluation fall outside this measure.

\field{Merging \(P_\ell\)}Write the attention output projection as \(W_O z+b_O\) and the following map as \(P_\ell(h)=A_\ell h+a_\ell\). After training, one affine map with weights \(A_\ell W_O\) and bias \(A_\ell b_O+a_\ell\) gives the same output. The separate global projection \(E_\ell(c_t)\) is evaluated once per sample per block and broadcast over image positions. On XL/2 the resulting generator has 701.75M parameters versus 676.89M for REG (Table~\ref{tab:params}). A single forward pass at batch size one costs 238.3 vs.\ 238.2 GFLOPs. At batch size 125 the median of 30 FP32 forward passes is 0.681\,s vs.\ 0.663\,s, a 2.7\% increase.

\field{XL/2 training measurements}Table~\ref{tab:xlcost} reports 1,000-update windows. All runs use a global batch of 256 and 16 images per device per microbatch. The eight-device runs accumulate two microbatches per update. The sixteen-device continuation to 2M uses one.

\begin{center}\begin{minipage}{\linewidth}
\captionof{table}{\textbf{Observed SiT-XL/2 training cost.} Time per successful update includes any skipped attempts between the endpoints. Memory is peak allocated tensor memory reported by rank zero up to the end of the window, in GiB.}
\label{tab:xlcost}\centering\small
\begin{tabular*}{\linewidth}{@{\extracolsep{\fill}}lrrrr@{}}
\toprule Run & Logical NPUs & Update interval & ms/update & Peak GiB\\\midrule
REG & 8 & 50,100--51,100 & 602.1 & 17.88\\
SSR & 8 & 50,100--51,100 & 593.8 & 25.37\\
SSR continuation & 16 & 200,100--201,100 & 377.1 & 22.58\\\bottomrule
\end{tabular*}\end{minipage}\end{center}

On eight devices, REG skipped no attempts and SSR skipped one. A single window on each of two different hosts cannot establish a speed advantage from the 8.3\,ms difference. The 7.49\,GiB increase is 41.9\% of the REG peak. SSR adds 62.05M trainable parameters, mostly in \(E_\ell\) and the unmerged \(P_\ell\), and stores local-state activations. The rank-zero peak includes the EMA copy and was recorded without a per-component breakdown.

The first 200K SSR updates used 259.62 logical NPU-hours on eight devices. The continuation used 3,126.48 on sixteen devices, totaling 3,386.10 through 2M, including 9,878 updates recomputed after a restart. These training-loop totals exclude downtime and offline evaluation.

\field{XL/2 checkpoint budgets}The ratios in Table~\ref{tab:budgets} compare training updates at evaluated checkpoints. Their wall-clock equivalents depend on per-step throughput and device count.
\begin{center}
\begin{minipage}{.96\linewidth}
\captionof{table}{\textbf{Measured checkpoints at published reference quality levels.} Unguided SiT-XL/2, ImageNet \(256\times256\). Hours are cumulative logical 910C NPU-hours for SSR. Ratios compare update counts.}
\label{tab:budgets}
\centering\small
\begin{tabular*}{\linewidth}{@{\extracolsep{\fill}}lrrrrrr@{}}
\toprule
& \multicolumn{2}{c}{Published reference} & \multicolumn{3}{c}{SSR checkpoint} & Update ratio\\
\cmidrule(lr){2-3}\cmidrule(lr){4-6}
Reference model & Steps & FID & Steps & FID & NPU-hours & \\
\midrule
SiT~\cite{supp-sit} & 7M & 8.3 & 70K & 6.84 & 91.40 & \(100\times\)\\
REPA~\cite{supp-repa} & 4M & 5.9 & 100K & 4.93 & 130.22 & \(40\times\)\\
REG~\cite{supp-reg} & 4M & 1.8 & 2M & 1.78 & 3,386.10 & \(2\times\)\\
\bottomrule
\end{tabular*}
\end{minipage}
\end{center}

\clearpage
\section{Extended ablations and guidance}
\label{sec:modules}
\subsection{Local-state modeling and global reading}

REG~\cite{supp-reg} jointly denoises the image latent and global CLS. The global switch gives every block direct access to the noisy CLS. The local switch adds the PCA patch state to denoising and reads it at the input (Table~\ref{tab:switches}).

\begin{center}
\begin{minipage}{\linewidth}
\captionof{table}{\textbf{Which paths are active?} SiT-B/2, ImageNet \(256\times256\), unguided \Fifty. All four configurations share the training seed and alignment weight \(\lambda=1\). The local state uses both input projections in Eq.~(2) of the main paper. Table~3 of that paper reports the separate REG + ReDi baseline~\cite{supp-redi}.}
\label{tab:switches}
\centering\small
\begin{tabular*}{\linewidth}{@{\extracolsep{\fill}}lccrrr@{}}
\toprule
& & & \multicolumn{3}{c}{Training iteration} \\
\cmidrule(lr){4-6}
Configuration & Local state & Direct global read & 100K & 200K & 400K\\
\midrule
REG & off & off & 32.16 & 20.22 & 14.35\\
REG + global read & off & on & 27.72 & 17.32 & 12.12\\
REG + local state & on & off & 26.99 & 16.62 & 11.46\\
SSR & on & on & \textbf{21.95} & \textbf{13.74} & \textbf{9.79}\\
\bottomrule
\end{tabular*}
\end{minipage}
\end{center}

\field{Conditional gains}

Let \(F_0,F_L,F_G,F_{LG}\) denote the FIDs for REG, local only, global only and full SSR. Adding global reading when the local state is already active reduces FID by 5.04 points at 100K and 1.67 at 400K (Table~\ref{tab:gains}).

\begin{center}
\begin{minipage}{.96\linewidth}
\captionof{table}{\textbf{Conditional reductions in FID.} Positive entries in the first four rows indicate lower FID after adding the module. The last row subtracts the sum of the standalone gains from the joint gain. All entries are arithmetic differences computed from Table~\ref{tab:switches}.}
\label{tab:gains}
\centering\small
\begin{tabular*}{\linewidth}{@{\extracolsep{\fill}}llrrr@{}}
\toprule
Change & Difference & 100K & 200K & 400K\\
\midrule
Add local state to REG & \(F_0-F_L\) & 5.17 & 3.60 & 2.89\\
Add local state with global read active & \(F_G-F_{LG}\) & 5.77 & 3.58 & 2.33\\
Add global read to REG & \(F_0-F_G\) & 4.44 & 2.90 & 2.23\\
Add global read with local state active & \(F_L-F_{LG}\) & 5.04 & 2.88 & 1.67\\
\midrule
Interaction on the FID scale & \(F_L+F_G-F_0-F_{LG}\) & \(+0.60\) & \(-0.02\) & \(-0.56\)\\
\bottomrule
\end{tabular*}
\end{minipage}
\end{center}

Each module improves FID with the other already active. The interaction changes sign across budgets and is within 0.02 FID of additivity at 200K.

\subsection{At which blocks, and from which state?}

With the local module active at 100K, using only the input CLS path gives FID 26.99. Adding a direct global read at the first block gives 24.54, at alternating blocks 22.64, and at every block 21.95. Holding all-layer reading fixed, FID for the current diffusion state \(c_t\) and the preceding block's normalized hidden CLS row is 40.22 vs.\ 41.80 at 50K, 21.95 vs.\ 22.82 at 100K and 13.74 vs.\ 13.80 at 200K. The FID gap narrows across these shared-seed checkpoints. At 200K the hidden CLS gives the lower sFID (6.59 vs.\ 6.74). The hidden-CLS run ended at 200K. The 400K results in Table~\ref{tab:switches} vary whether the local and global paths are active. They leave the relative performance of the two read sources at 400K open.

At 200K, sharing \(E,P\), sharing \(E\) and removing \(P\), or merging \(P_\ell\) before training gives FIDs 14.61, 14.35 and 14.16. Separate per-block maps give 13.74 (Table~3 of the main paper).

\clearpage
\subsection{Guiding three denoising predictions}
\label{sec:evaluation}

Let \(d_{u,\mathrm{cond}}\) and \(d_{u,\mathrm{uncond}}\) be the conditional and unconditional drifts for \(u\in\{x,c,s\}\). Within the guidance interval, we apply classifier-free guidance (CFG)~\cite{supp-cfg} to each component as \(d_{u,\mathrm{uncond}}+w_u(d_{u,\mathrm{cond}}-d_{u,\mathrm{uncond}})\). Outside that interval the CFG correction is zero. Equation~\eqref{eq:score} gives \(d_u=u_t+(2-t)v_u\). At fixed \(u_t,t\), the guidance weights sum to one, so applying them to velocities before computing the drift gives the same update as combining drifts.
\begin{center}
\begin{minipage}{\linewidth}
\captionof{table}{\textbf{Guidance sensitivity at one checkpoint.} Every row uses the same SiT-B/2 SSR 400K EMA weights, 50,000 samples, seed convention and 250-step sampler. CFG is active at evaluation times $0<t\le t_{\max}$. A patch strength of one uses the ordinary class-conditional local prediction.}
\label{tab:sweepb}
\centering\small
\begin{tabular*}{\linewidth}{@{\extracolsep{\fill}}rrrrrrrr@{}}
\toprule $w_x=w_c$ & $w_s$ & $t_{\max}$ & FID $\downarrow$ & sFID $\downarrow$ & IS $\uparrow$ & Prec. $\uparrow$ & Rec. $\uparrow$\\
\midrule
1.3 & 1.3 & 1.00 & 5.11 & 5.69 & 183.6 & 0.772 & 0.565 \\
1.5 & 1.5 & 1.00 & 4.12 & 5.53 & 220.5 & 0.801 & 0.526 \\
1.8 & 1.8 & 1.00 & 3.93 & 5.48 & 266.7 & 0.830 & 0.482 \\
2.0 & 2.0 & 1.00 & 4.22 & 5.51 & 289.7 & 0.841 & 0.459 \\
1.5 & 1.5 & 0.85 & 6.87 & 5.99 & 155.0 & 0.726 & 0.605 \\
2.3 & 1.0 & 0.85 & 5.79 & 5.90 & 176.0 & 0.730 & 0.604 \\
2.3 & 2.3 & 0.85 & 4.44 & 5.73 & 207.4 & 0.747 & 0.592 \\
\bottomrule
\end{tabular*}
\end{minipage}
\end{center}

For B/2, full-interval CFG at 1.8 gives the lowest measured FID of 3.93 (Table~\ref{tab:sweepb}). Holding the strength at 1.5 and shortening the interval raises FID from 4.12 to 6.87. Within the shorter interval, keeping \(w_x=w_c=2.3\) and raising \(w_s\) from 1 to 2.3 lowers FID from 5.79 to 4.44. For the XL/2 pure-CFG comparison, we carried over the image and CLS settings \(w=2.3,\ t_{\max}=0.85\) from an earlier REG-based sampler, used the same weight for the local stream and held all three fixed across checkpoints.

\begin{center}
\begin{minipage}{.98\linewidth}
\captionof{table}{\textbf{Change in XL/2 metrics at fixed training budgets.} Each entry subtracts pure CFG from CFG with autoguidance using the reported values in Table~2 of the main paper. Both settings use the same EMA checkpoint, 50,000 samples and 250 sampling steps.}
\label{tab:supp-guided}
\centering\small
\begin{tabular*}{\linewidth}{@{\extracolsep{\fill}}lrrrrr@{}}
\toprule
Epochs & \(\Delta\)FID & \(\Delta\)sFID & \(\Delta\)IS & \(\Delta\)Precision & \(\Delta\)Recall\\
\midrule
80  & \(-0.04\) & \(+0.44\) & \(-22.9\) & \(-0.01\) & \(+0.02\)\\
160 & \(-0.09\) & \(+0.60\) & \(-19.5\) & \(-0.01\) & \(+0.01\)\\
200 & \(-0.08\) & \(+0.61\) & \(-20.4\) & \(-0.01\) & \(+0.02\)\\
\bottomrule
\end{tabular*}
\end{minipage}
\end{center}

Pure CFG on XL/2 reaches FID 1.34 at 400 epochs. For the hybrid sampler, let \(d_u^+\) and \(d_u^\varnothing\) denote the XL/2 conditional and unconditional drifts. The weak drift \(d_u^{\rm weak}\) comes from the 100K-step SiT-L/2 SSR EMA evaluated at the same joint state, time and class label as \(d_u^+\). At every positive grid time we use
\[
 d_u^{\rm hyb}=d_u^+
 +0.8\,\mathbf{1}_{\{t\leq0.85\}}\bigl(d_u^+-d_u^\varnothing\bigr)
 +0.5\bigl(d_u^+-d_u^{\rm weak}\bigr),\qquad u\in\{x,c,s\}.
\]
Thus CFG has scale 1.8 inside its interval. Autoguidance~\cite{supp-autoguidance} remains active above and below the cutoff on all three streams. The final deterministic step uses only the image drift. We selected this setting by FID at 400K updates and reused it at 800K and 1M. For the 1M evaluation, four ranks sampled batches of 125 each, using FP32 model and VAE evaluations and FP64 integration. Rank \(r=0,\ldots,3\) used CPU seed \(r\) for the image, label and CLS draws and \(777+r\) for local noise. At 1M (200 epochs), FID falls from 1.39 to 1.31 and recall rises from 0.63 to 0.65, while sFID rises from 4.12 to 4.73 (Table~\ref{tab:supp-guided}). Both the CFG scale and the additional model change in this comparison. Measured hybrid sampling time is \(1.39\times\) that of pure CFG.

\clearpage
\section{Probing the semantic states}
\label{sec:mechanism}
\subsection{Recovering semantics from partially corrupted images}
\label{sec:recovery}

We use the 400K-step SiT-B/2 EMA checkpoints for REG and SSR from the \(\lambda=1\) ablation in Table~\ref{tab:switches}. Following REG~\cite{supp-reg}, we test on 10,000 ImageNet validation crops, ten per class. The two models share the corrupted image, class label, Gaussian CLS draw and subsequent image and CLS noise. SSR uses independent noise for its local state. Frozen DINOv2-B~\cite{supp-dinov2} supplies the clean scoring targets.

\field{Recovery protocol}At $t=0.498795$ the image is a linear interpolation of the scaled SD-VAE posterior mode of the $256\times256$ center crop and Gaussian noise. The semantic states start from pure Gaussian noise at this same time, with no clean semantic component. We run the remaining unguided steps of the 250-point grid and read $\hat u_0=u_t-t\,v_u$ at $t=0.04$.

\begin{center}
\begin{minipage}{\linewidth}
\captionof{table}{\textbf{Image-specific CLS recovery.} Each predicted CLS is scored against the DINOv2-B CLS of its input. The different-image control uses the next validation image of the same class in cyclic order. Brackets give paired 95\% bootstrap intervals over the 1,000 classes.}
\label{tab:recovery}
\centering\small
\begin{tabular*}{\linewidth}{@{\extracolsep{\fill}}lrrr@{}}
\toprule Measure & REG & SSR & SSR $-$ REG\\
\midrule
CLS cosine, corresponding image & 0.0494 & 0.0383 & \(-0.0111\) [\(-0.0117, -0.0103\)] \\
CLS cosine, different image of the same class & 0.0418 & 0.0365 & \(-0.0053\) [\(-0.0060, -0.0046\)] \\
Corresponding minus different image & 0.0076 & 0.0018 & \(-0.0058\) [\(-0.0061, -0.0054\)] \\
Neighborhood alignment (CKNNA, $k=10$) & 0.00277 & 0.00177 & \(-0.00100\) \\
\bottomrule
\end{tabular*}
\end{minipage}
\end{center}

\field{Global and local comparisons}The small cosines in Table~\ref{tab:recovery} compare a recovered CLS with a clean target under this initialization. Scoring two clean CLS vectors for the same cyclic within-class pairs gives mean cosine 0.430. CKNNA~\cite{supp-platonic} uses ten neighbors per unit-normalized CLS among all 10,000 crops, excluding the query. We score each recovered patch against its corresponding whitened PCA target, then circularly shift the reference grid by eight cells on both axes as a control.

Across the 10,000 SSR reconstructions, mean patch cosine is 0.184 at corresponding positions and \(-0.120\) after shifting. The paired difference is +0.303 (95\% class-bootstrap interval [+0.302, +0.305]). The intervals use 2,000 resamples with fixed model weights.

\field{Retrieving an image among ten of the same class}We rank ten clean candidates by CLS cosine, mean cosine of 256 corresponding patches, or their prespecified equal-weight mean (Table~\ref{tab:retrieval}). Chance is 10\%. The rerun that saved the grids reproduced the original CLS vectors and both patch scores for every image.

\begin{center}\begin{minipage}{\linewidth}
\captionof{table}{\textbf{Within-class instance retrieval under partial corruption.} Top-1 accuracy on 10,000 queries. Brackets give 95\% class-bootstrap intervals. The last column gives the paired difference from REG CLS in percentage points.}
\label{tab:retrieval}\centering\small
\begin{tabular*}{\linewidth}{@{\extracolsep{\fill}}llrr@{}}
\toprule Model & Scoring feature & Top-1 accuracy (\%) & Difference from REG CLS (pp)\\\midrule
REG & Global CLS & 14.27 [13.55, 14.97] & Reference\\
SSR & Global CLS & 10.74 [10.13, 11.35] & \(-3.53\) [\(-4.12,-2.96\)]\\
SSR & Local patch grid & \textbf{35.56} [34.64, 36.48] & \(+21.29\) [20.12, 22.45]\\
SSR & Equal-weight CLS and grid & 23.76 [22.88, 24.70] & \(+9.49\) [8.60, 10.36]\\
\bottomrule\end{tabular*}\end{minipage}\end{center}

Adding the CLS score overturns 1,978 correct patch-only matches and recovers 798 misses, a net loss of 11.80 percentage points (paired 95\% interval \([-12.83,-10.80]\)). All ten candidates share a class label, so this task rewards features that distinguish individual instances.

\field{Class information in the recovered CLS}In each of ten folds we hold out one image per class in image-ID order and rank 1,000 prototypes by cosine. Each prototype averages the nine other unit-normalized clean CLS vectors of its class. Over the same 10,000 queries, top-1 accuracy is 75.92\% using the clean query CLS, 4.92\% using recovered REG CLS and 2.77\% using recovered SSR CLS (chance 0.10\%). This weak above-chance class signal is compatible with a coarse-context role. Both models receive the true label, and this probe does not isolate the effect of initialization or explain the FID gains from per-layer reading in Table~\ref{tab:gains}.

\clearpage
\subsection{Intervening on the generated semantic trajectory}
\label{sec:trajectory}
We sample trajectories A and B with different labels and noise. At the first solver grid point at or below the specified \(t\), we replace A's local state \(s_t\), global state \(c_t\), or both with B's corresponding state. Each transferred stream uses B's subsequent noise. A retains its label, its image trajectory up to the swap and its subsequent image noise. Any semantic stream that was kept continues with A's noise.

\begin{center}\begin{minipage}{\linewidth}
\centering\includegraphics[width=\linewidth]{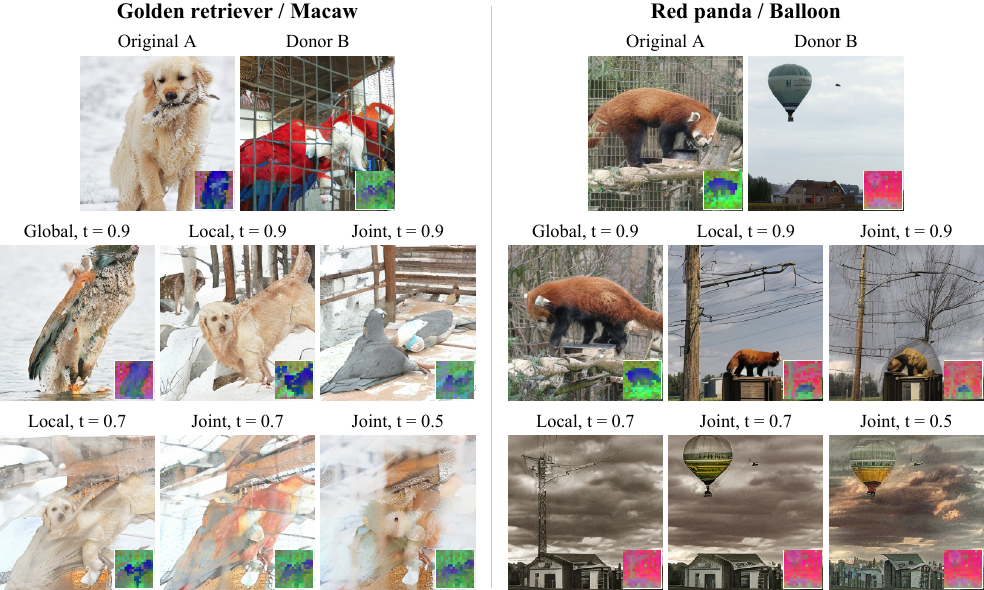}\par
\captionof{figure}{\textbf{Global, local and joint swaps.} The two class pairs here complement those in Fig.~\ref{fig:swap}. Each intervention starts from A and transfers the indicated state from B. Insets share one three-component projection of the final local state. The 2M-step XL/2 EMA uses 250 unguided evaluations with seed 4. Table~\ref{tab:displacement} reports seeds 0--3.}
\label{fig:swaps}\end{minipage}\end{center}

\begin{center}\begin{minipage}{\linewidth}
\captionof{table}{\textbf{Displacement of the final image latent.} Each cell reports the mean and sample standard deviation over four class pairs and four seeds (16 observations).}
\label{tab:displacement}\centering\small
\begin{tabular*}{.9\linewidth}{@{\extracolsep{\fill}}lccc@{}}
\toprule Transferred state & \(t=0.9\) & \(t=0.7\) & \(t=0.5\)\\\midrule
Local \(s\) & \(0.548\pm 0.080\) & \(0.433\pm 0.044\) & \(0.313\pm 0.039\)\\
Global \(c\) & \(0.364\pm 0.090\) & \(0.226\pm 0.046\) & \(0.130\pm 0.028\)\\
Joint \(c,s\) & \(0.577\pm 0.077\) & \(0.442\pm 0.047\) & \(0.323\pm 0.039\)\\
\bottomrule\end{tabular*}\end{minipage}\end{center}
In Figure~\ref{fig:swaps}, the red panda stays on a branch after the global-only swap at \(t=0.9\). The local-only swap at the same time places it on a box under overhead wires. With a joint swap at \(t=0.7\), a balloon appears above the rooftops.

Table~\ref{tab:displacement} reports distances from A in the final image latent, normalized by \(D=\|x_B-x_A\|_2\) for each pair and seed. Additional runs reproduced every A and B endpoint exactly. Local-only swaps move farther from A than global-only swaps at all three times. At \(t=0.9,0.7,0.5\), adding the global swap to the local swap increases mean distance from A by only 0.029, 0.009 and 0.010. Yet the mean distance \emph{between} those two endpoints, \(\|x_{c,s}-x_s\|_2/D\), is 0.441, 0.266 and 0.152, respectively. Changing the global trajectory still changes the final image latent even when its distance from A barely changes. These distances measure neither class identity nor a specific visual attribute.

\clearpage
\section{Additional qualitative results}
\label{sec:samples}
\subsection{Training checkpoints under matched image noise}

Figure~\ref{fig:paired} extends the checkpoint comparison in Fig.~\ref{fig:compare} to two additional ImageNet classes, African elephant and school bus. Each class uses identical initial image noise and noise increments across all nine method--checkpoint combinations. Sampling uses SiT-XL/2 EMA weights, 250 SDE steps and CFG 4.0 throughout on every generated variable. At 50K, the SSR elephant has a visible eye, tusk and hanging trunk. The trunk is less distinct in both baseline samples. The SSR school bus shows a row of windows above two visible wheels, while the baseline views show smaller sections of the vehicle. Table~\ref{tab:supp-guided} compares two other XL/2 guidance settings at matched checkpoints.

\begin{center}
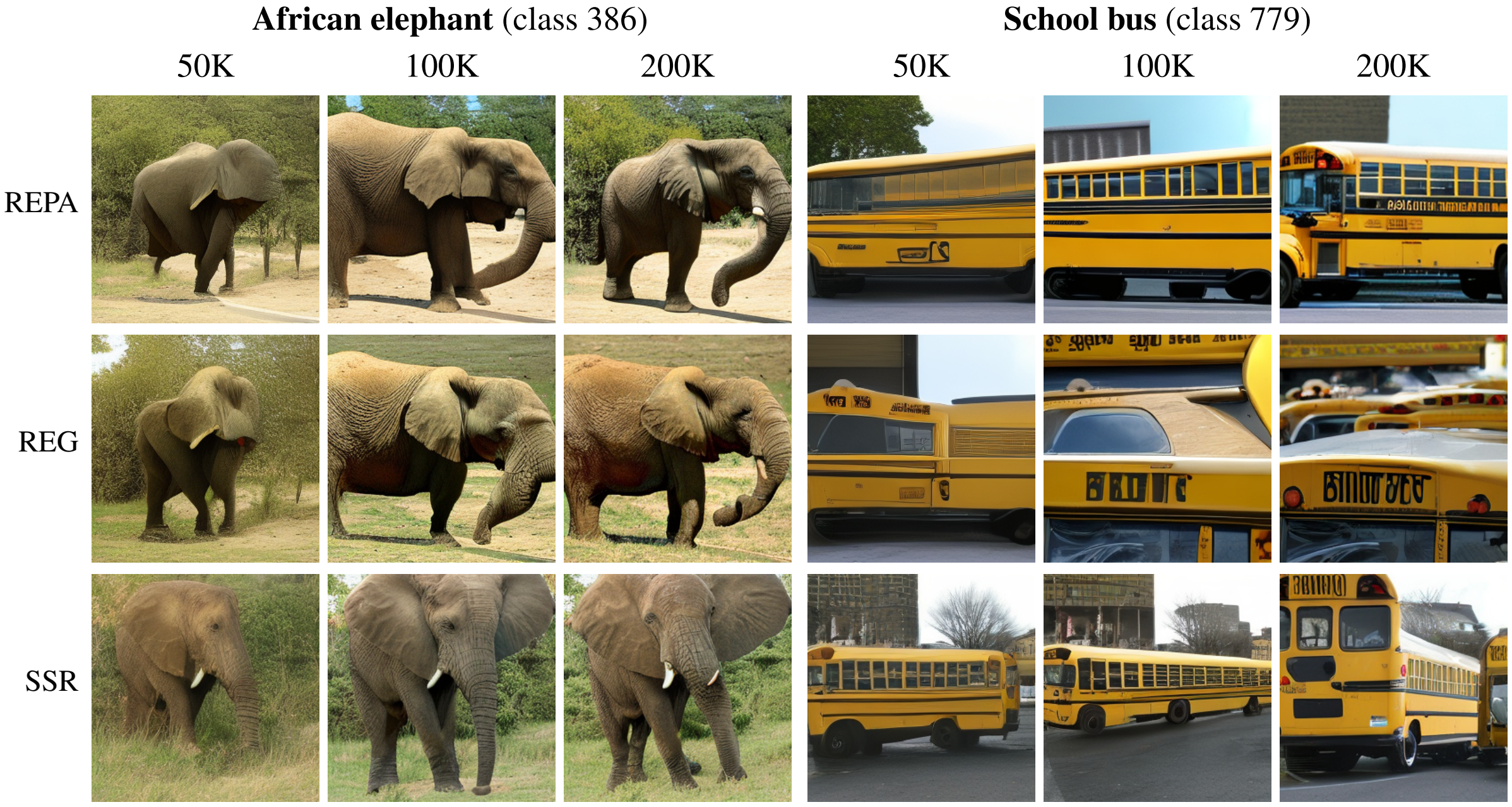

\begin{minipage}{\linewidth}
\centering
\begingroup
\setlength{\tabcolsep}{1.3pt}
\newcommand{\tile}[3]{\includegraphics[width=.142\linewidth]{compare/#1-#2-#3.png}}
\begin{tabular}{@{}r@{\hspace{4pt}}ccc@{\hspace{5pt}}ccc@{}}
& \multicolumn{3}{c}{\textbf{African elephant} (class 386)} & \multicolumn{3}{c}{\textbf{School bus} (class 779)}\\[2pt]
& 50K & 100K & 200K & 50K & 100K & 200K\\[2pt]
\raisebox{.069\linewidth}{\small REPA} & \tile{repa}{50000}{386} & \tile{repa}{100000}{386} & \tile{repa}{200000}{386} & \tile{repa}{50000}{779} & \tile{repa}{100000}{779} & \tile{repa}{200000}{779}\\
\raisebox{.069\linewidth}{\small REG} & \tile{reg}{50000}{386} & \tile{reg}{100000}{386} & \tile{reg}{200000}{386} & \tile{reg}{50000}{779} & \tile{reg}{100000}{779} & \tile{reg}{200000}{779}\\
\raisebox{.069\linewidth}{\small SSR} & \tile{ssr}{50000}{386} & \tile{ssr}{100000}{386} & \tile{ssr}{200000}{386} & \tile{ssr}{50000}{779} & \tile{ssr}{100000}{779} & \tile{ssr}{200000}{779}\\
\end{tabular}
\endgroup
\captionof{figure}{\textbf{Selected examples at 50K, 100K and 200K training iterations.} We show two of 32 matched class--noise pairs (16 classes, two draws each), selected for clearer subject structure in early SSR samples. The image seeds are 100773 (left) and 101558 (right). The CLS and local streams, when present, use seed offsets of 10,000 and 20,000.}
\label{fig:paired}
\end{minipage}
\end{center}

\subsection{Paired noise with and without guidance}
\label{sec:gallery}
Figures~\ref{fig:gallery1}--\ref{fig:gallery4} show five of sixteen draws for each of twelve ImageNet classes. We selected the 60 pairs from the 192 generated pairs for subject visibility and variety of composition. The draw numbers are the original within-class indices. Each pair uses the same initial Gaussian states and subsequent noise increments for all three streams, the 2M-step XL/2 EMA checkpoint, and 250 sampling evaluations. The lower row applies \(w_x=w_c=w_s=2.3\) for \(t\leq0.85\).

In the selected pairs, matched noise can preserve the framing of a subject while guidance changes its color, pose or the number of objects. The selection illustrates these differences without estimating their frequency. Figure~\ref{fig:all} shows all sixteen draws for the first and last classes in the fixed list, in draw order. For volcano draw 4 the ash plume occupies much the same part of the frame in both images. In goldfish draw 7 the number and placement of visible hands change. Figure~\ref{fig:limitations} enlarges three cases where a foreground subject or its surroundings are difficult to resolve.

\clearpage
\begingroup
\newcommand{\ssrGalleryPage}[3]{\noindent\begin{minipage}{\linewidth}
  \centering\includegraphics[width=\linewidth]{#1}\par
  \captionof{figure}{#2}\label{#3}
  \end{minipage}\par\clearpage}

\ssrGalleryPage{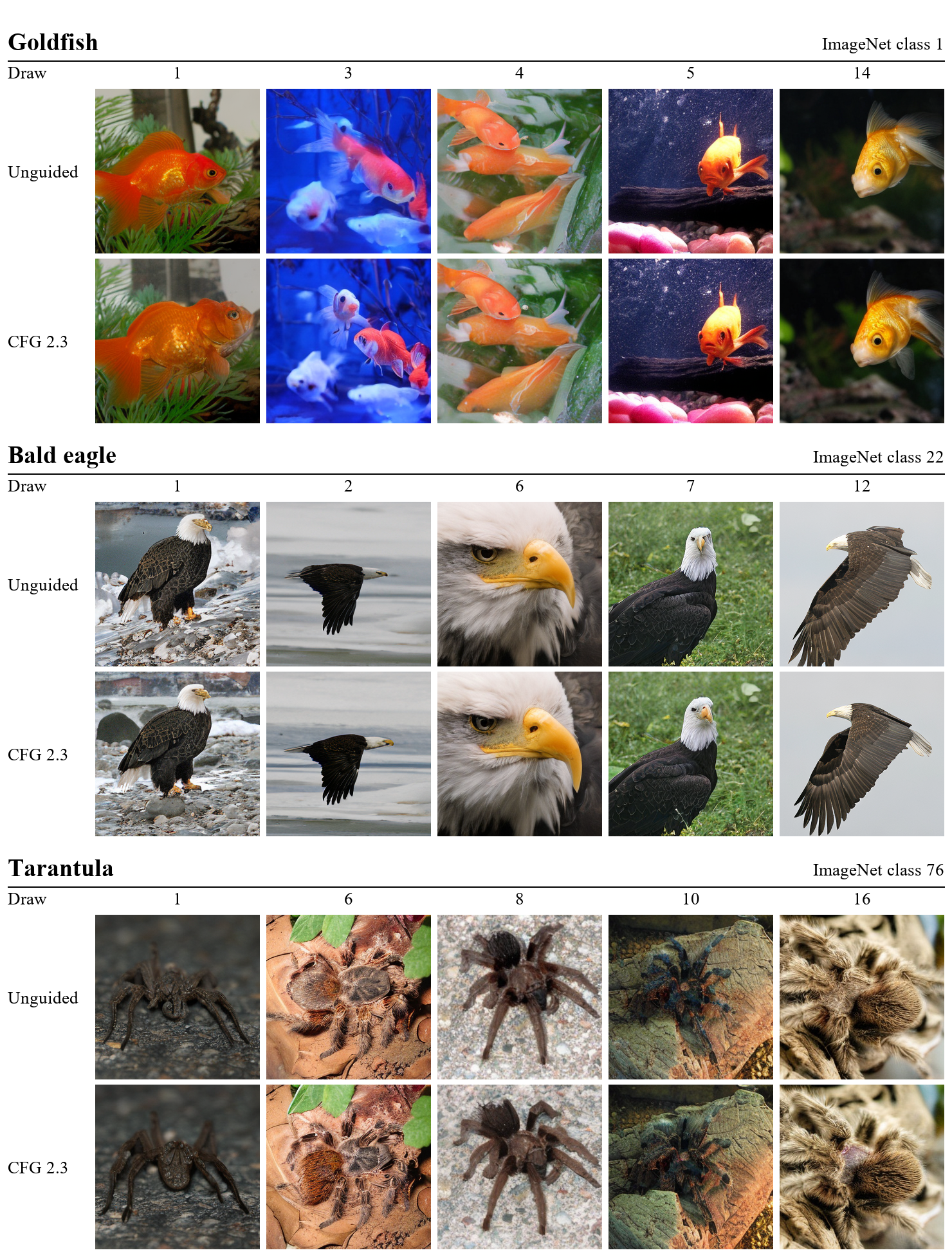}{\textbf{Goldfish, bald eagle and tarantula.} Bald eagle draw 6 retains the close-up of the eye and hooked beak under both settings. Goldfish draw 3 retains the blue aquarium, while the arrangement of the fish changes.}{fig:gallery1}
\ssrGalleryPage{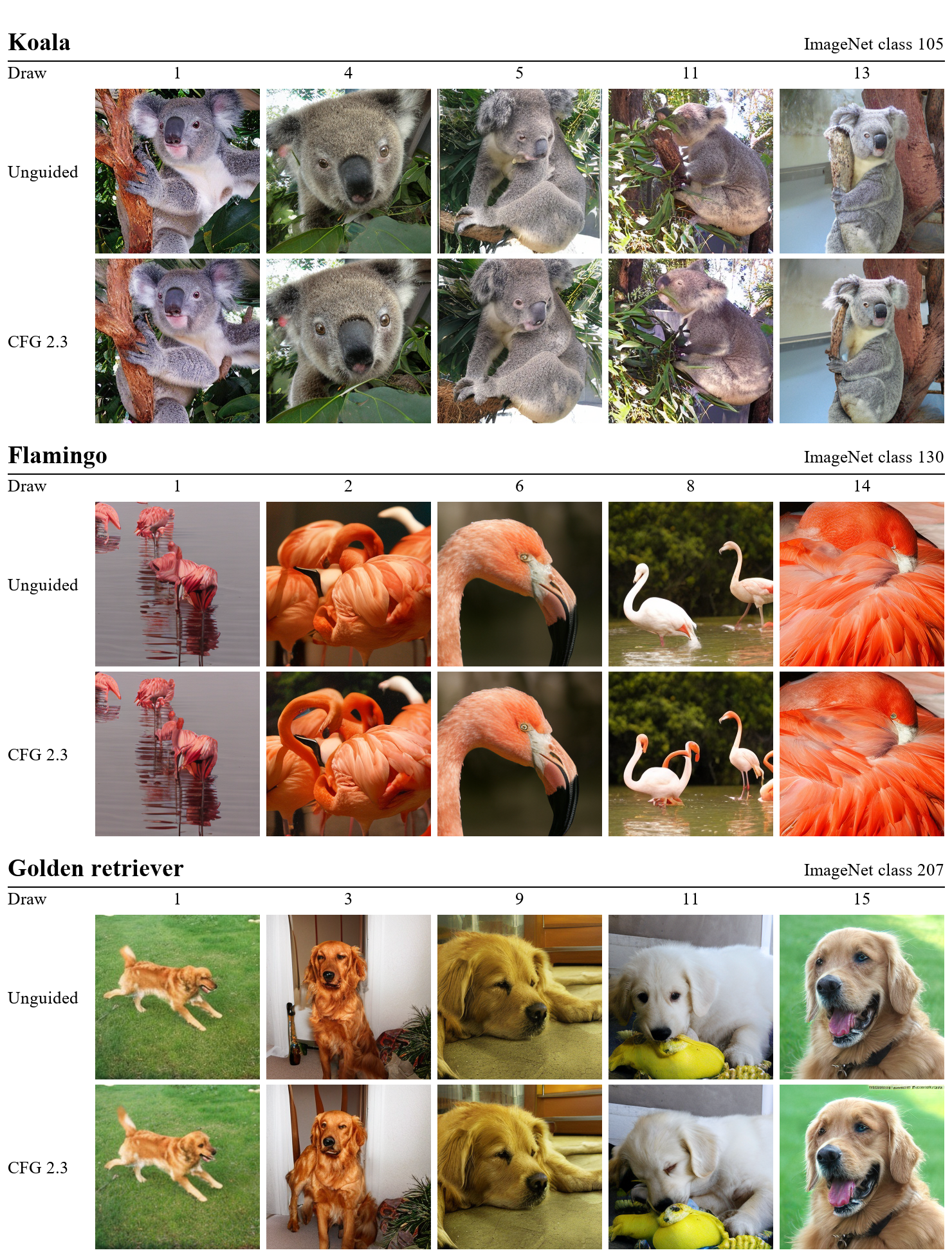}{\textbf{Koala, flamingo and golden retriever.} In koala draw 1 the animal clings to the same forked tree in both rows. In flamingo draw 8 the waterline persists, while the number and spacing of the birds change.}{fig:gallery2}
\ssrGalleryPage{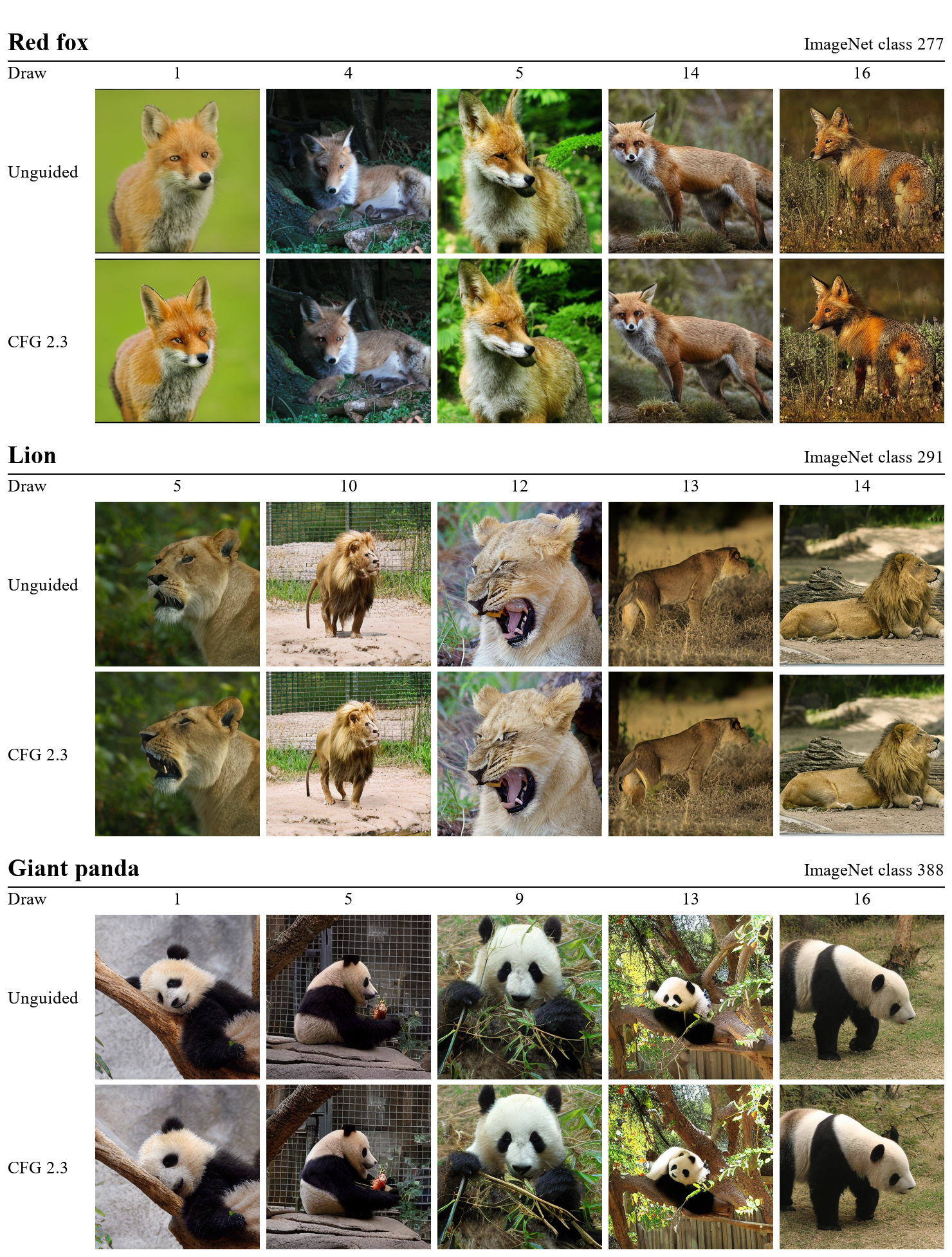}{\textbf{Red fox, lion and giant panda.} Red fox draw 1 retains the upright pose against green ground, with more saturated orange fur under CFG. In lion draw 5 the head remains in profile, but the mouth opens wider under guidance.}{fig:gallery3}
\ssrGalleryPage{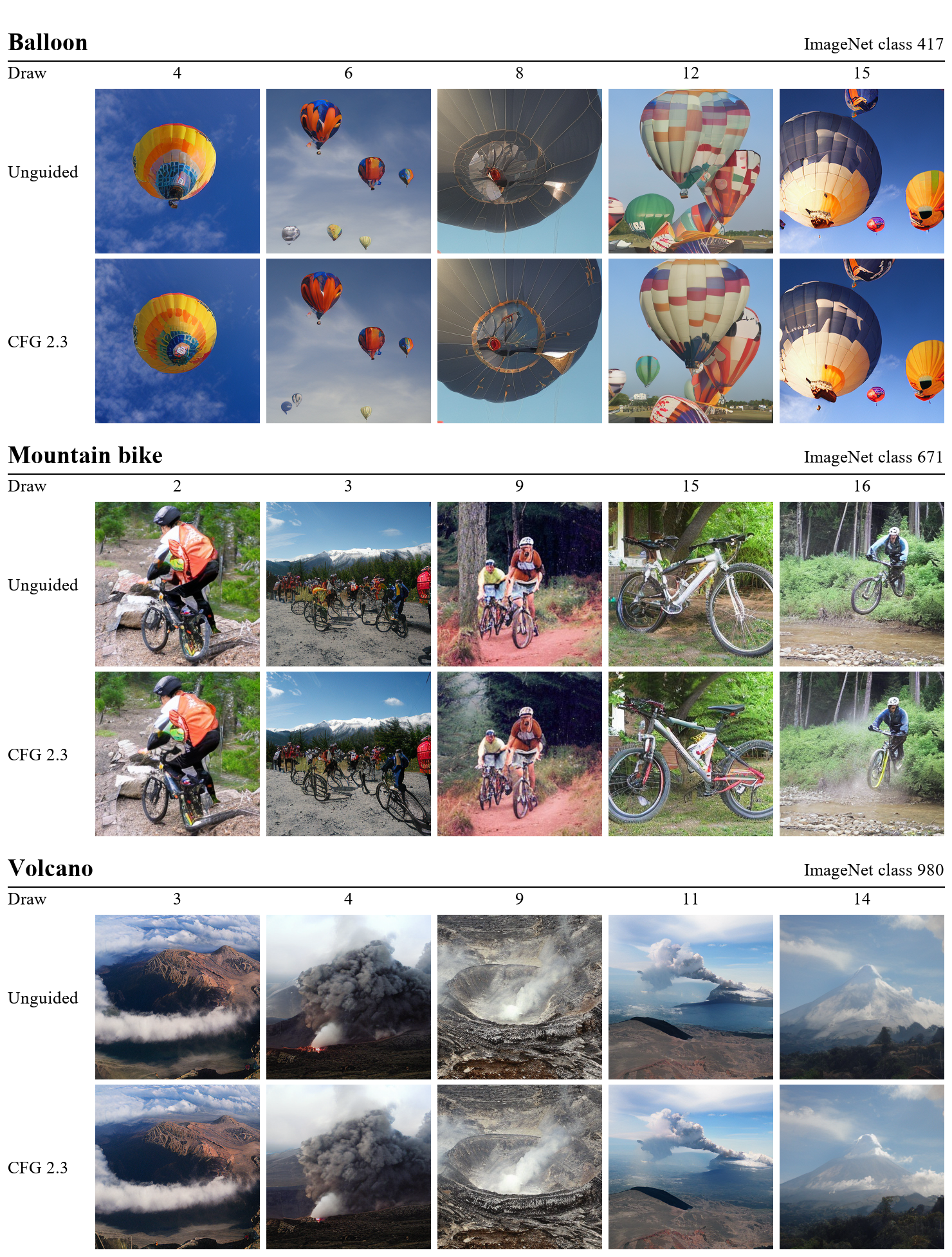}{\textbf{Balloon, mountain bike and volcano.} In mountain bike draw 2 the cyclist leans over the same rocky step in both rows. Volcano draw 4 retains the dark plume rising above the vent under both settings.}{fig:gallery4}
\endgroup
\subsection{All draws for two classes}
We show the first and last classes in the fixed twelve-class list. Each class contributes all sixteen consecutive draws. The upper and lower rows of each group share the noise realization for the indicated draw.\par
\begingroup\setlength{\parskip}{0pt}\setlength{\tabcolsep}{0pt}
\vspace{5pt}
{\large\bfseries Goldfish}\hfill ImageNet class 1\par\vspace{2pt}\hrule\vspace{4pt}
\begin{tabular*}{\linewidth}{@{\extracolsep{\fill}}l*{8}{c}@{}}
\small Draw & \small 1 & \small 2 & \small 3 & \small 4 & \small 5 & \small 6 & \small 7 & \small 8\\[1pt]
\raisebox{.044\linewidth}{\small Unguided} & \includegraphics[width=.109\linewidth]{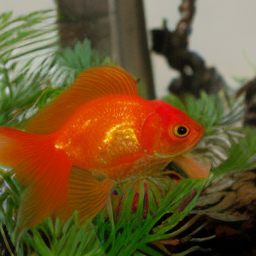} & \includegraphics[width=.109\linewidth]{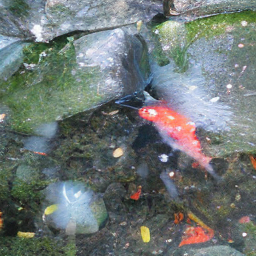} & \includegraphics[width=.109\linewidth]{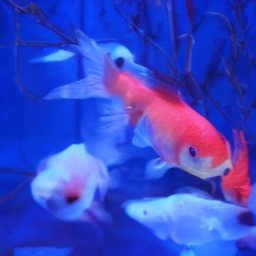} & \includegraphics[width=.109\linewidth]{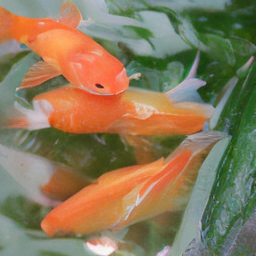} & \includegraphics[width=.109\linewidth]{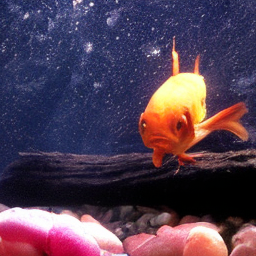} & \includegraphics[width=.109\linewidth]{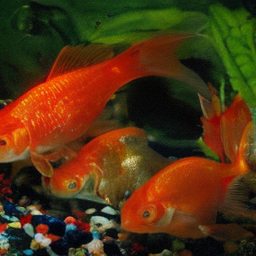} & \includegraphics[width=.109\linewidth]{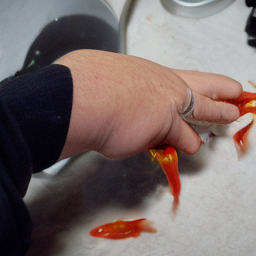} & \includegraphics[width=.109\linewidth]{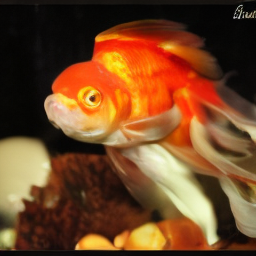}\\
\raisebox{.044\linewidth}{\small CFG 2.3} & \includegraphics[width=.109\linewidth]{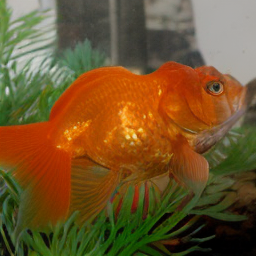} & \includegraphics[width=.109\linewidth]{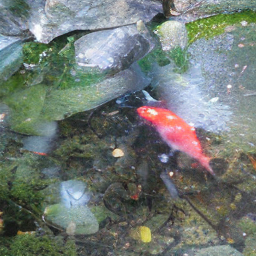} & \includegraphics[width=.109\linewidth]{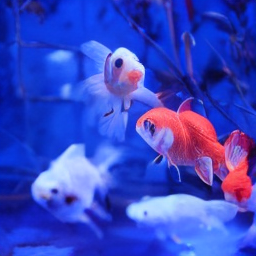} & \includegraphics[width=.109\linewidth]{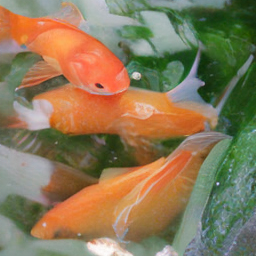} & \includegraphics[width=.109\linewidth]{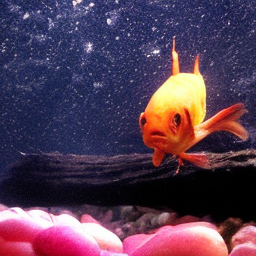} & \includegraphics[width=.109\linewidth]{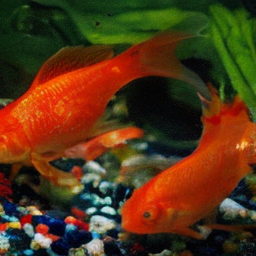} & \includegraphics[width=.109\linewidth]{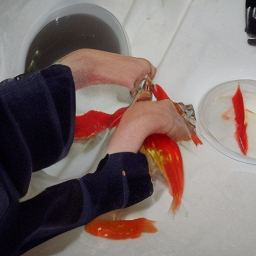} & \includegraphics[width=.109\linewidth]{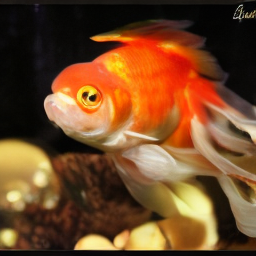}\\
\end{tabular*}\par
\vskip4pt plus 1fill
\begin{tabular*}{\linewidth}{@{\extracolsep{\fill}}l*{8}{c}@{}}
\small Draw & \small 9 & \small 10 & \small 11 & \small 12 & \small 13 & \small 14 & \small 15 & \small 16\\[1pt]
\raisebox{.044\linewidth}{\small Unguided} & \includegraphics[width=.109\linewidth]{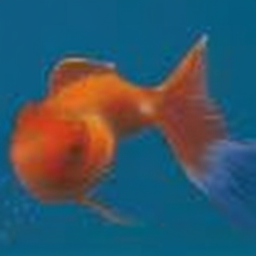} & \includegraphics[width=.109\linewidth]{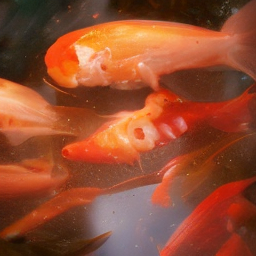} & \includegraphics[width=.109\linewidth]{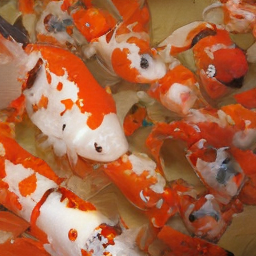} & \includegraphics[width=.109\linewidth]{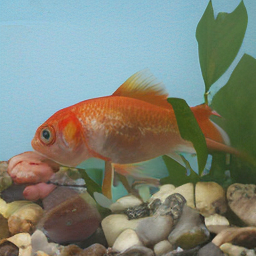} & \includegraphics[width=.109\linewidth]{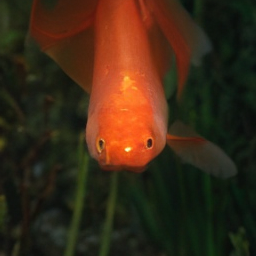} & \includegraphics[width=.109\linewidth]{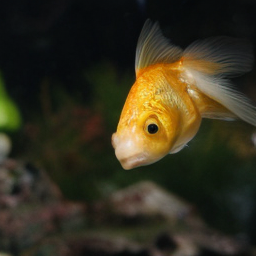} & \includegraphics[width=.109\linewidth]{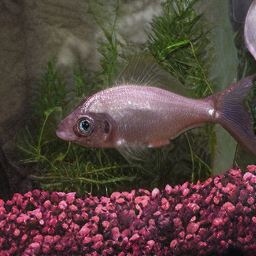} & \includegraphics[width=.109\linewidth]{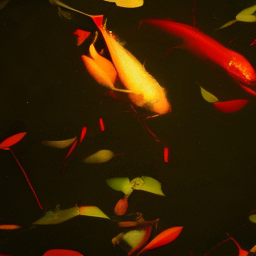}\\
\raisebox{.044\linewidth}{\small CFG 2.3} & \includegraphics[width=.109\linewidth]{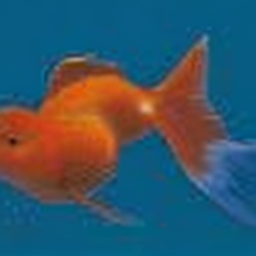} & \includegraphics[width=.109\linewidth]{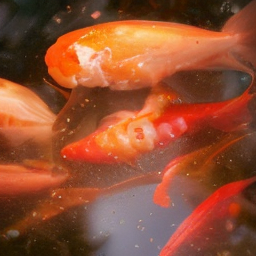} & \includegraphics[width=.109\linewidth]{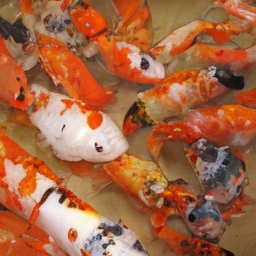} & \includegraphics[width=.109\linewidth]{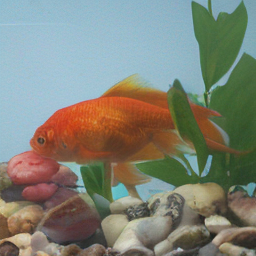} & \includegraphics[width=.109\linewidth]{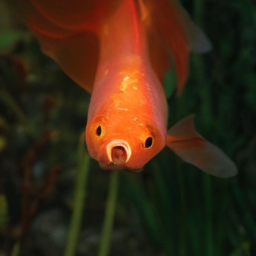} & \includegraphics[width=.109\linewidth]{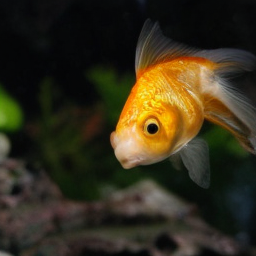} & \includegraphics[width=.109\linewidth]{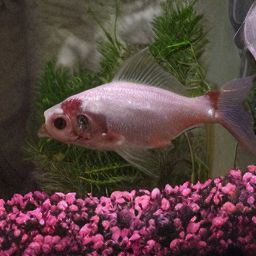} & \includegraphics[width=.109\linewidth]{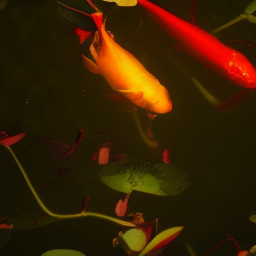}\\
\end{tabular*}\par
\vskip6pt plus 1fill
\vspace{5pt}
{\large\bfseries Volcano}\hfill ImageNet class 980\par\vspace{2pt}\hrule\vspace{4pt}
\begin{tabular*}{\linewidth}{@{\extracolsep{\fill}}l*{8}{c}@{}}
\small Draw & \small 1 & \small 2 & \small 3 & \small 4 & \small 5 & \small 6 & \small 7 & \small 8\\[1pt]
\raisebox{.044\linewidth}{\small Unguided} & \includegraphics[width=.109\linewidth]{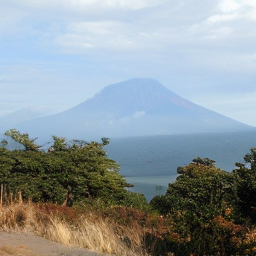} & \includegraphics[width=.109\linewidth]{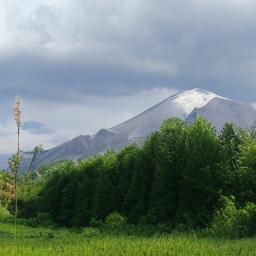} & \includegraphics[width=.109\linewidth]{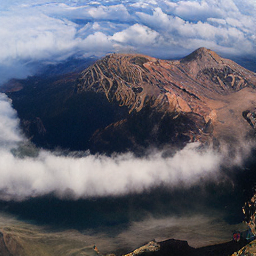} & \includegraphics[width=.109\linewidth]{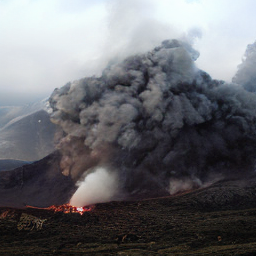} & \includegraphics[width=.109\linewidth]{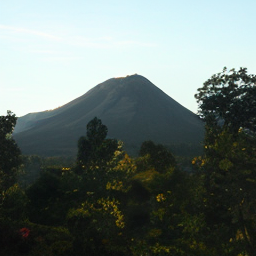} & \includegraphics[width=.109\linewidth]{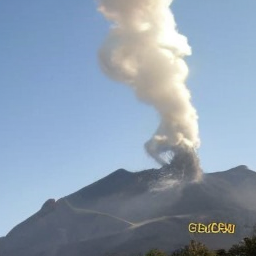} & \includegraphics[width=.109\linewidth]{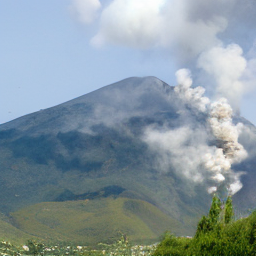} & \includegraphics[width=.109\linewidth]{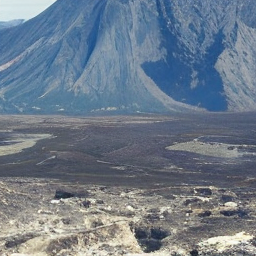}\\
\raisebox{.044\linewidth}{\small CFG 2.3} & \includegraphics[width=.109\linewidth]{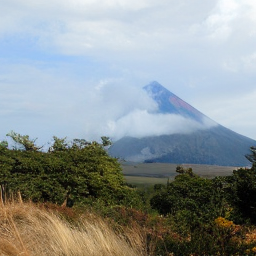} & \includegraphics[width=.109\linewidth]{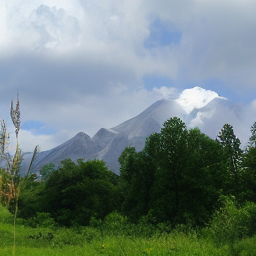} & \includegraphics[width=.109\linewidth]{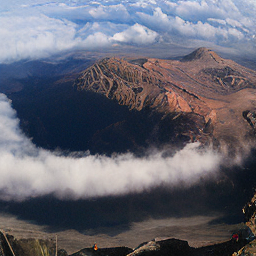} & \includegraphics[width=.109\linewidth]{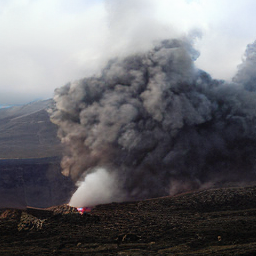} & \includegraphics[width=.109\linewidth]{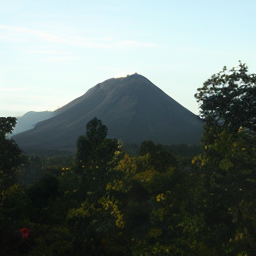} & \includegraphics[width=.109\linewidth]{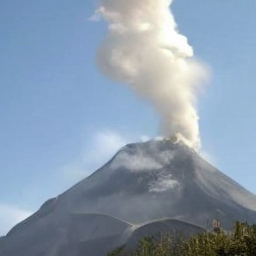} & \includegraphics[width=.109\linewidth]{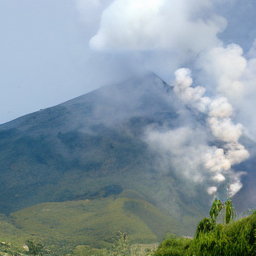} & \includegraphics[width=.109\linewidth]{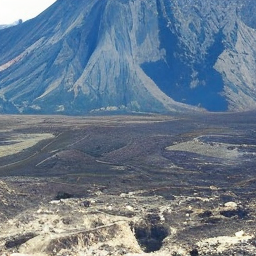}\\
\end{tabular*}\par
\vskip4pt plus 1fill
\begin{tabular*}{\linewidth}{@{\extracolsep{\fill}}l*{8}{c}@{}}
\small Draw & \small 9 & \small 10 & \small 11 & \small 12 & \small 13 & \small 14 & \small 15 & \small 16\\[1pt]
\raisebox{.044\linewidth}{\small Unguided} & \includegraphics[width=.109\linewidth]{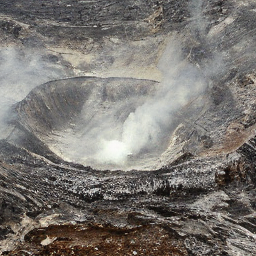} & \includegraphics[width=.109\linewidth]{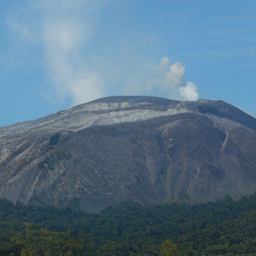} & \includegraphics[width=.109\linewidth]{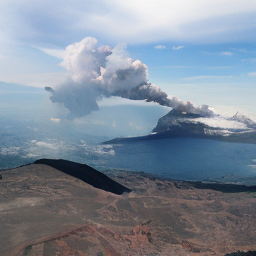} & \includegraphics[width=.109\linewidth]{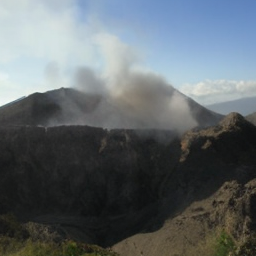} & \includegraphics[width=.109\linewidth]{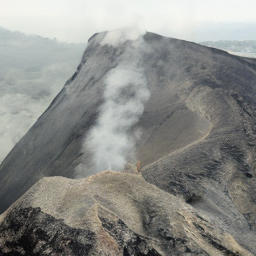} & \includegraphics[width=.109\linewidth]{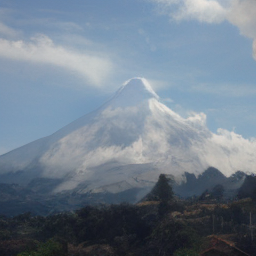} & \includegraphics[width=.109\linewidth]{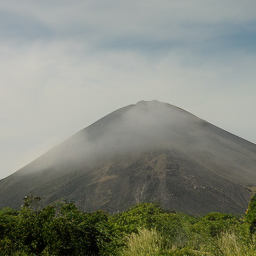} & \includegraphics[width=.109\linewidth]{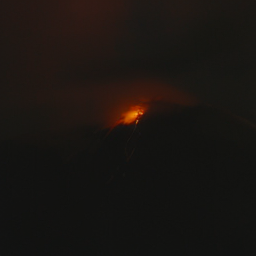}\\
\raisebox{.044\linewidth}{\small CFG 2.3} & \includegraphics[width=.109\linewidth]{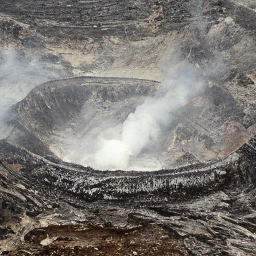} & \includegraphics[width=.109\linewidth]{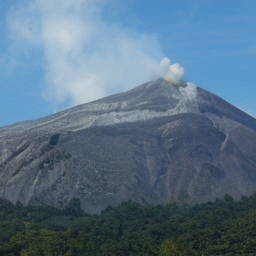} & \includegraphics[width=.109\linewidth]{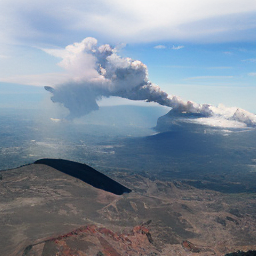} & \includegraphics[width=.109\linewidth]{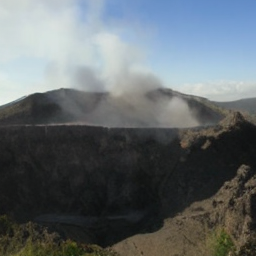} & \includegraphics[width=.109\linewidth]{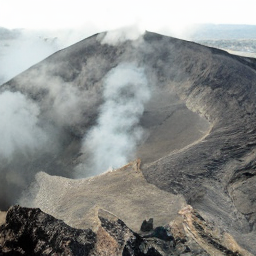} & \includegraphics[width=.109\linewidth]{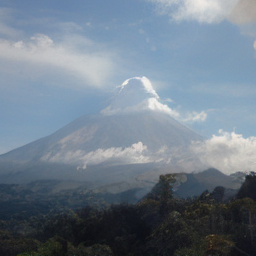} & \includegraphics[width=.109\linewidth]{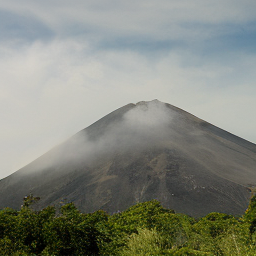} & \includegraphics[width=.109\linewidth]{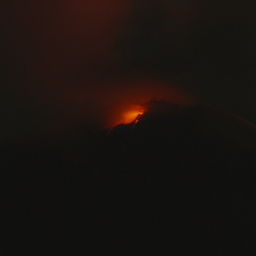}\\
\end{tabular*}\par
\endgroup
\vspace{6pt}
\begin{minipage}{\linewidth}
\captionof{figure}{\textbf{Uncurated paired samples.} All 32 pairs are shown in draw order. Goldfish draw 9 remains soft in both rows. Volcano draw 4 preserves the ash plume under guidance, while draw 16 has little visible detail outside the glow.}
\label{fig:all}
\end{minipage}
\clearpage
\subsection{Difficult examples}
\begin{center}\begin{minipage}{\linewidth}\centering
\begingroup\setlength{\tabcolsep}{3pt}\begin{tabular}{@{}lccc@{}}
 & Goldfish, draw 9 & Mountain bike, draw 5 & Volcano, draw 16\\[3pt]
\raisebox{.108\linewidth}{\small Unguided} & \includegraphics[width=.245\linewidth]{pairs/class_001-draw-9-unguided.png} & \includegraphics[width=.245\linewidth]{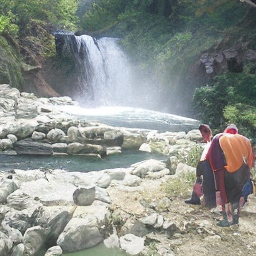} & \includegraphics[width=.245\linewidth]{pairs/class_980-draw-16-unguided.png}\\
\raisebox{.108\linewidth}{\small CFG 2.3} & \includegraphics[width=.245\linewidth]{pairs/class_001-draw-9-cfg23.png} & \includegraphics[width=.245\linewidth]{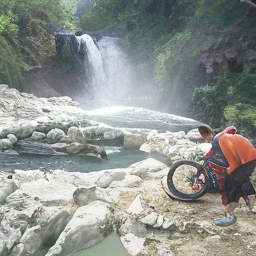} & \includegraphics[width=.245\linewidth]{pairs/class_980-draw-16-cfg23.png}\\
\end{tabular}\endgroup
\captionof{figure}{\textbf{Three difficult draws from the same sample pool.} The goldfish has indistinct fins in both images. In the mountain-bike pair the waterfall dominates the unguided image, and a bicycle wheel becomes visible near the right edge under CFG. In the volcano pair the scene is nearly dark except for a small glow. These examples illustrate specific limitations. Their selection does not measure how often each occurs.}
\label{fig:limitations}
\end{minipage}\end{center}

\end{document}